%% file: main.tex
\documentclass{article}
\usepackage[final]{neurips_2021}

\usepackage[utf8]{inputenc} 
\usepackage[T1]{fontenc}    
\usepackage{hyperref}       
\usepackage{url}            
\usepackage{booktabs}       
\usepackage{amsfonts}       
\usepackage{nicefrac}       
\usepackage{microtype}      
\usepackage{xcolor}         
\usepackage{graphicx}
\usepackage{wrapfig}
\usepackage{xcolor}

\title{Can You Learn an Algorithm?  Generalizing from Easy to Hard Problems with Recurrent Networks}

\author{
  Avi Schwarzschild\\
  Department of Mathematics\\
  University of Maryland\\
  College Park, MD, USA\\
  \texttt{avi1@umd.edu}\\
  \And
  Eitan Borgnia\\
  Department of Computer Science\\
  University of Maryland\\
  College Park, MD, USA\\
  \And
  Arjun Gupta\\
  Department of Robotics\\
  University of Maryland\\
  College Park, MD, USA\\
  \And
  Furong Huang\\
  Department of Computer Science\\
  University of Maryland\\
  College Park, MD, USA\\
  \And
  Uzi Vishkin\\
  Department of Electrical and Computer Engineering\\
  University of Maryland\\
  College Park, MD, USA\\
  \And
  Micah Goldblum\\
  Department of Computer Science\\
  University of Maryland\\
  College Park, MD, USA\\
  \And
  Tom Goldstein\\
  Department of Computer Science\\
  University of Maryland\\
  College Park, MD, USA\\
}

\begin{document}

\maketitle


\begin{abstract}
    Deep neural networks are powerful machines for visual pattern recognition, but reasoning tasks that are easy for humans may still be difficult for neural models. Humans possess the ability to extrapolate reasoning strategies learned on simple problems to solve harder examples, often by thinking for longer. For example, a person who has learned to solve small mazes can easily extend the very same search techniques to solve much larger mazes by spending more time.  In computers, this behavior is often achieved through the use of algorithms, which scale to arbitrarily hard problem instances at the cost of more computation. In contrast, the sequential computing budget of feed-forward neural networks is limited by their depth, and networks trained on simple problems have no way of extending their reasoning to accommodate harder problems. In this work, we show that recurrent networks trained to solve simple problems with few recurrent steps can indeed solve much more complex problems simply by performing additional recurrences during inference. We demonstrate this algorithmic behavior of recurrent networks on prefix sum computation, mazes, and chess.  In all three domains, networks trained on simple problem instances are able to extend their reasoning abilities at test time simply by ``thinking for longer.''
\end{abstract}

\section{Introduction}
\label{sec:intro}

In computational theories of mind, an analytical problem is tackled by embedding it in ``working memory'' and then iteratively applying transformations to the representation until the problem is solved \citep{baddeley2012working,baddeley1974working}.  Iterative processes underlie the human ability to solve sequential reasoning problems, such as complex question answering, proof writing, and even object classification \citep{liao2016bridging, kar2019evidence}.  They also enable humans to extrapolate their knowledge to solve problems of potentially unbounded complexity, including harder problems than they have seen before, by thinking for longer.

This work examines whether recurrent neural networks trained on easy problems can extrapolate their knowledge to solve hard problems. 
We find that recurrent networks can indeed generalize to harder problems simply by increasing their test time iteration budget (i.e., thinking for longer than they did at train time). Moreover, we find that the performance of recurrent models improves as they recur for more iterations, even without adding parameters or re-training in the new, more challenging problem domain. This ability is specific to recurrent networks, as standard feed-forward networks rely on layer-specific behaviors that cannot be repeated to extend their reasoning power.  
 
The behavior we observe in recurrent networks falls outside the classical notions of generalization in which models are trained and tested on the same distribution.  Because we train and test on problems of different sizes/difficulties, our training and test distributions are disjoint, and systems must extrapolate to solve problems from the test distribution.  Outside the field of machine learning, computers achieve a functionally similar extrapolation ability through the use of algorithms, which encode the process required to solve a class of problems, and can therefore scale to problems of arbitrary size, albeit with longer runtime. 
 
By training networks to solve problems iteratively, we hope to find models that encode a scalable {\em method} for solving problems rather than {\em memorizing} a mapping between input features and outputs.  In short, the goal is to create recurrent architectures that are capable of {\em learning an algorithm}.


Our focus is on three reasoning problems that are classically solved using hand-crafted algorithms: computing prefix sums, solving mazes, and playing chess.  Sequential reasoning tasks like these are ideal for our study because one can directly quantify the difficulty of a problem instance. In the case of mazes, for example, we can easily swap to a more challenging domain by increasing the size of the search space.

For each class of problems, recurrent networks are trained on a set of ``easy’’ problems using a fixed number of iterations of the recurrent module on the forward pass.  After training is complete, we assess whether our models exhibit logical extrapolation behaviors by testing them on ``hard’’ problems, with varying numbers of additional iterations. Remarkably, models trained on easy examples exhibit little extrapolative behavior until their iteration budget is increased — generalizing to harder problems \emph{requires} thinking deeper. Moreover, we find recurrent models tested with a sufficient number of extra iterations outperform the inflexible feed-forward models of comparable depth, often by a wide margin. Finally, we visualize the iterative behavior of the recurrent module to gain insights into the problem solving process they discover.


\subsection{Related works}

Our investigation into generalizing from easy to hard examples builds on several bodies of work. Logical extrapolation encompasses a special kind of distributional shift.  A number of existing works on domain generalization instead explore shifts, such as re-stylization and image corruptions, which do not represent an increase in scale or computational complexity \citep{arjovsky2019invariant, shu2020preparing}. Also, the basic neural architectures we use are not new and build upon prior studies of weight sharing and recurrence \citep{pinheiro2014recurrent, liang2015recurrent, alom2018recurrent, bai2018trellis, bai2019deep, lan2020albert, jaegle2021perceiver}. Networks with variable numbers of test time iterations/layers have also been studied, including variable depth networks \citep{graves2016adaptive, huang2016deep, kaya2019shallow, eyzaguirre2020differentiable}.

Existing work on algorithm learning involves recurrent neural network (RNN) based approaches. For example, neural turing machines and neural GPUs can learn simple algorithms for tasks such as binary addition and multiplication \citep{graves2014neural, kaiser2015neural}. Like most RNNs, the compute budget for these methods is inextricably tied to input length. Motivated by the fact that input sequence length is not necessarily correlated with the computational burden required to solve a problem, \citet{graves2016adaptive} develops a method for RNNs to adaptively select a compute time limit. This work considers only sequence inputs and shows the benefits of decoupling compute budget from input length. A differentiable extension of the technique can also be applied to visual question answering \citep{eyzaguirre2020differentiable}. 

The above works leverage neural networks with adaptive computation budgets to speed up and strengthen inference when learning on stationary distributions. In contrast, our work studies the logical extrapolation behaviors that recurrent networks possess when both computation budgets and problem difficulties are extended beyond the train-time regime. In the domain of constraint satisfiability problems (CSPs), \cite{selsam2018learning} show message passing neural networks trained on small CSPs can generalize to larger problems if more messages are passed at test-time -- a similar type of extrapolation to our methods, but for a very specific problem formulation. 

The particular problems we use to study this type of extrapolation include prefix sum computation and maze solving, two problems analyzed in the classical algorithms literature. For example, there are many ways to solve mazes, both classical (e.g. breadth first search) and learned (e.g. value iteration networks), but our goal is not to develop the best solver, rather to use mazes as a test bed for logical extrapolation \citep{tamar2016value}. Our third case study is the game of chess, which has also been the focus of much artificial intelligence work \citep{stockfish, biswas2015measuring, silver2017mastering, mcilroy2020aligning}. However, those efforts to play chess rely heavily on hand-crafted search algorithms, often paired with neural networks or opening books, and aim to play games from start to finish, both methods and goals that diverge from ours. As opposed to hard-coded algorithms, which scale by design, we are interested in studying whether learned processes can generalize from the data on which they are trained to even harder problems.

\section{Dataset descriptions}
\label{sec:data}

We conduct experiments in three problem domains that are classically solved using hand-crafted algorithms. For each, we define datasets that have quantifiable notions of difficulty. This makes it possible to train models on easy/small examples and test them on harder/larger ones. We consider the task of computing prefix sums modulo two of binary bit strings, solving two-dimensional mazes, and finding the best move in chess puzzles.

\paragraph{\textbf{Prefix sums}}
This problem is inspired by a similar dataset used by \cite{graves2016adaptive}. 
Each training sample is a binary string. The goal is to output a binary string of equal length, where each bit represents the cumulative sum of input bits modulo two. Our models accept input strings of any size, and we consider longer strings to be more difficult to process than shorter ones. Each dataset contains 10,000 uniform random binary strings without duplicates. We use datasets with input lengths of 32, 44, and 48 bits. See Figure \ref{fig:prefix_example} for an example input and the corresponding target output.

\begin{figure}[h]
    \centering
    \begin{tabular}{cccccccccccccccccc}
        Input: & $[1, 0, 0, 1, 0, 1, 0, 1,  0, 1, 0, 1,1, 0, 1,1, 0, 1]$ \\
        Target: & $[1, 1, 1, 0, 0, 1, 1, 0, 0, 1, 1, 0, 1, 1, 0, 1, 1, 0]$
    \end{tabular}
    \caption{Prefix sum input and target example.}
    \label{fig:prefix_example}
\end{figure}


\paragraph{\textbf{Mazes}}
We consider mazes generated using a depth-first search algorithm \citep{hill2017making}. 
We train on 50,000 small ($9 \times 9$) mazes, and we test on 10,000 larger ($13 \times 13$) mazes. 
Our models are convolutional, and receive a maze as a $N \times N$ three-channel image, where the maze walls are black, and the start and goal locations are red and green, respectively. The label for each maze is a binary two-dimensional mask containing the locations of positions along the shortest path solution. 
Our models output a two logits per pixel which are then thresholded to a binary mask of the whole input, and we consider a candidate solution correct only if it exactly predicts the labeled path.
See Figure \ref{fig:maze_example} for an example. More examples are available in Appendix \ref{sec:app_maze_data}.

\begin{figure}[b]
\vspace{-12pt}
\centering
    \begin{minipage}[b]{0.45\textwidth}
    \centering
    \includegraphics[width=0.46\textwidth]{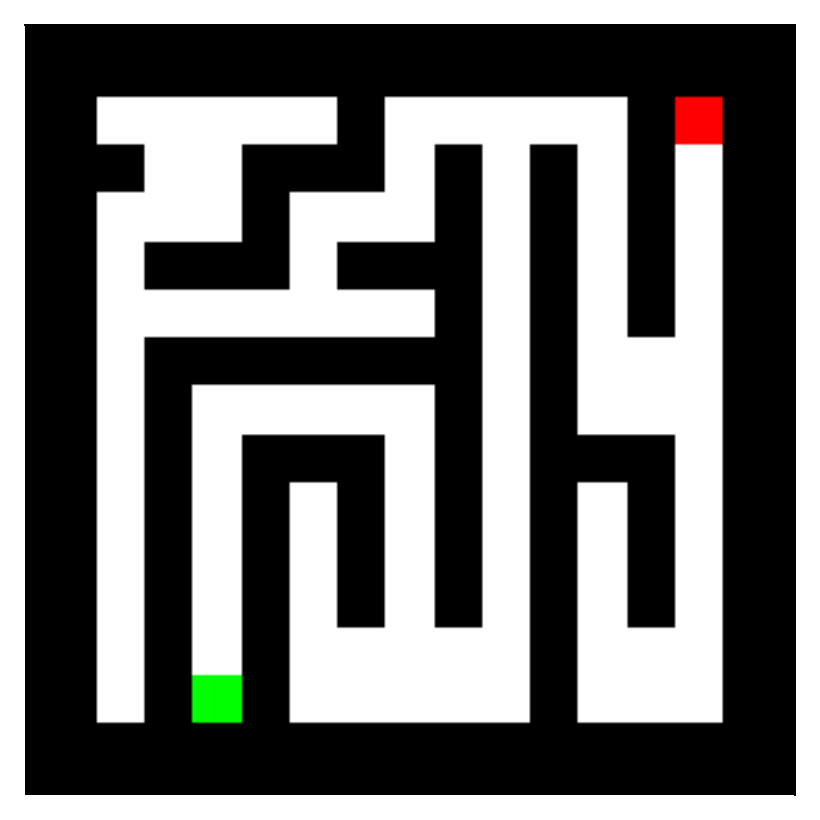}
    \includegraphics[width=0.475\textwidth, trim=1cm 1.1cm 1cm 1cm, clip]{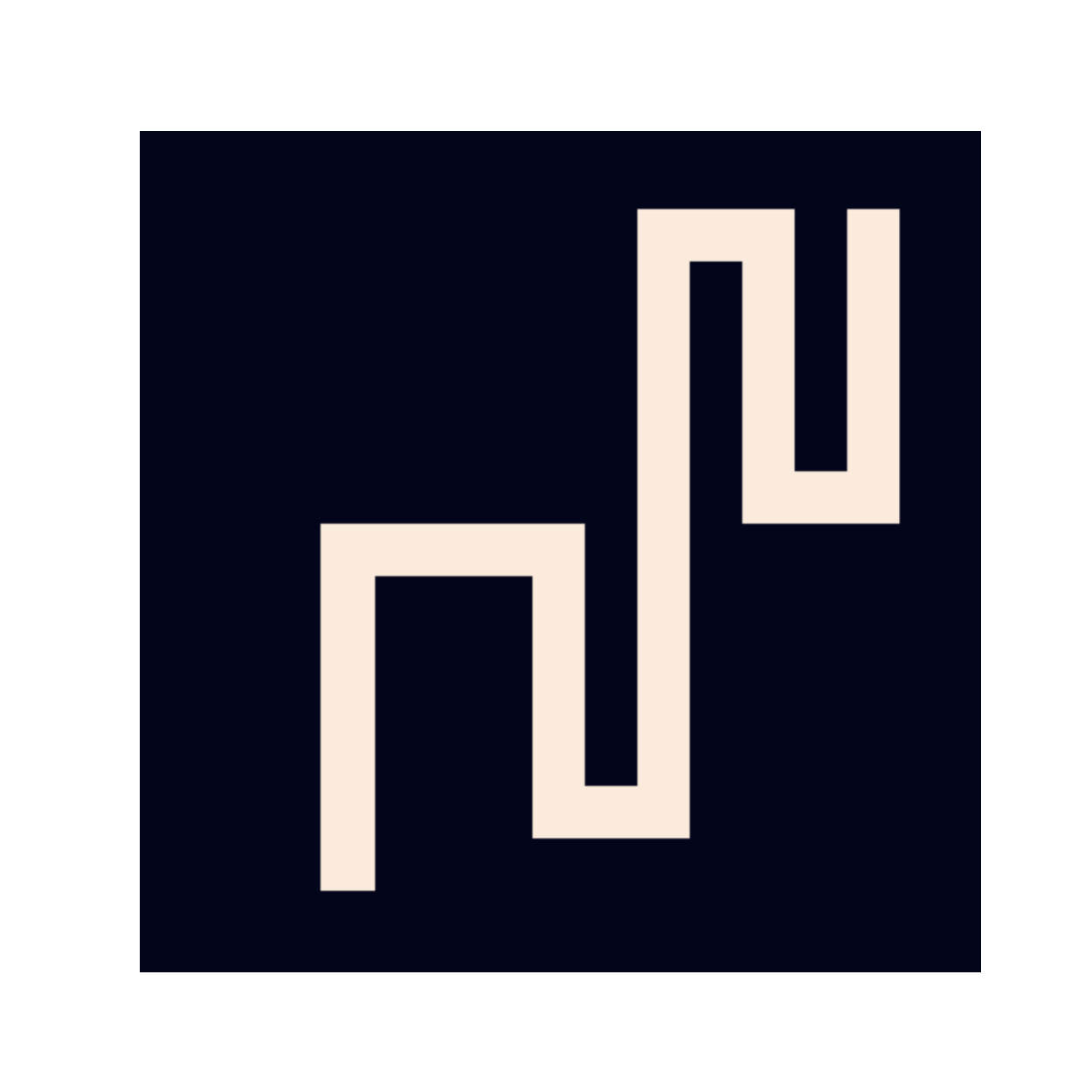}
    \caption{An example maze input and target. The target is a binary classification label for each pixel indicating on/off the optimal path.}
    \label{fig:maze_example}
    \end{minipage}
    \hspace{12pt}
    \begin{minipage}[b]{0.45\textwidth}
    \centering
    \includegraphics[width=0.46\textwidth]{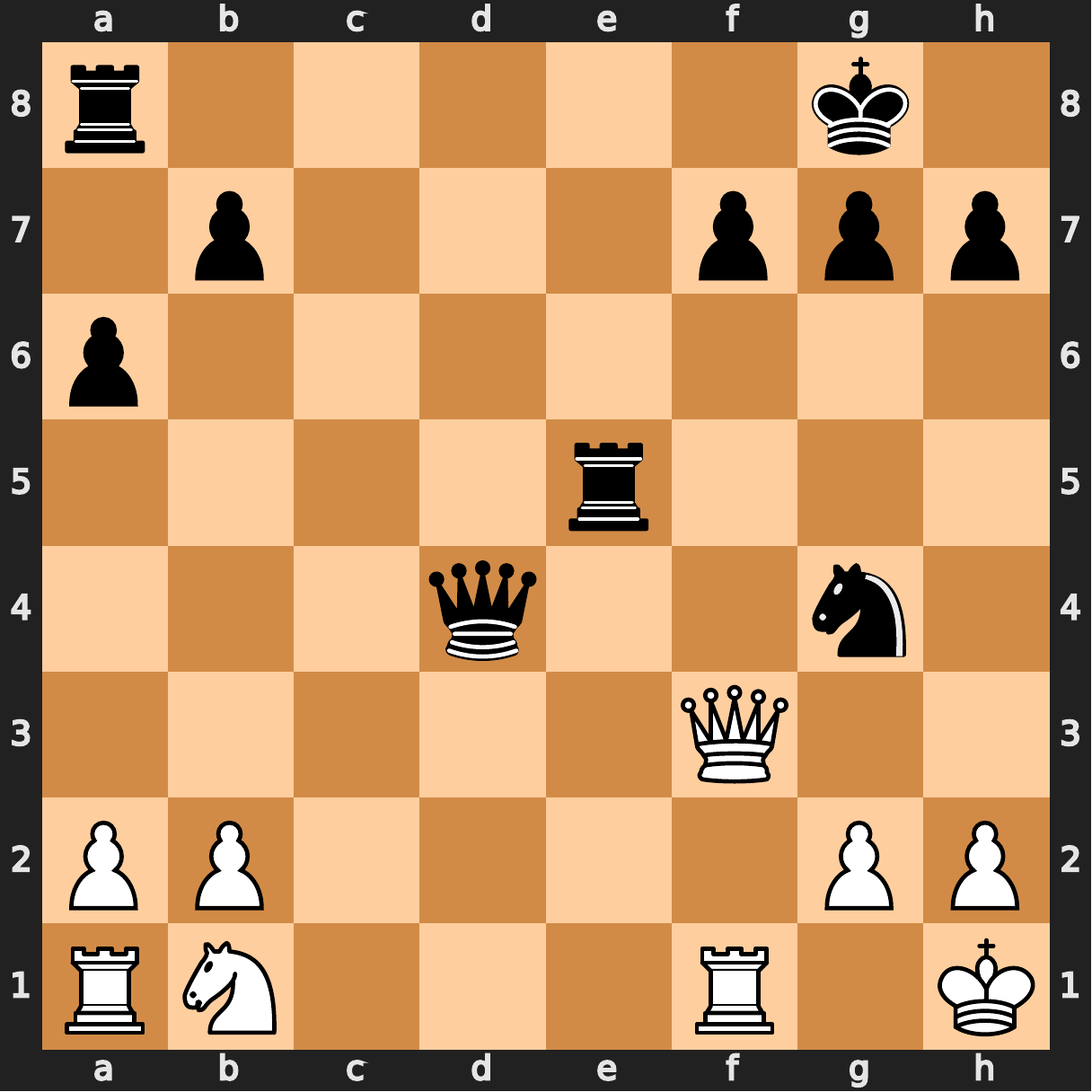}
    \includegraphics[width=0.48\textwidth, trim=1cm 1.1cm 1cm 1cm, clip]{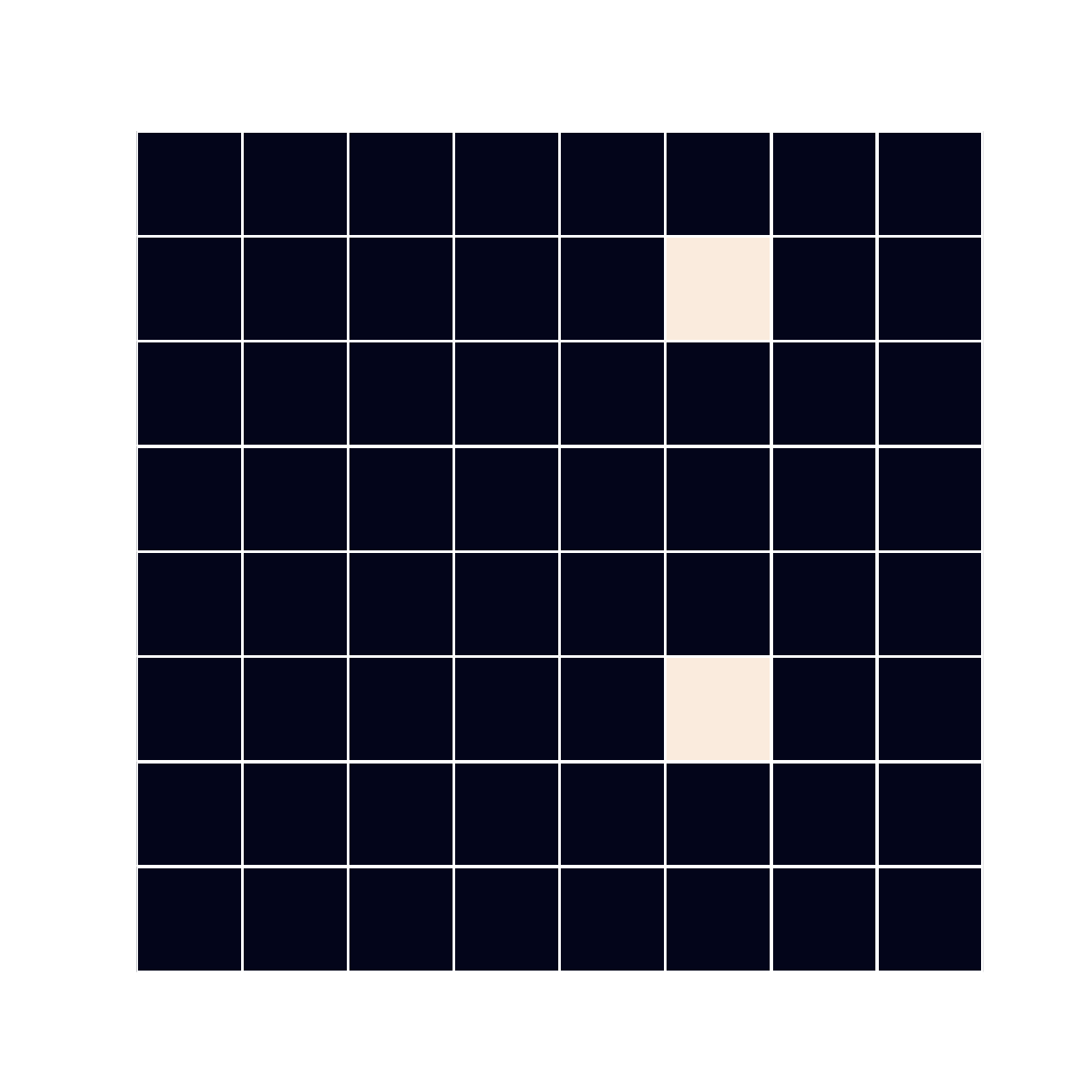}
    \caption{An example of a chess puzzle input (left) and target (right). In the target, white represents one and black is zero. }
    \label{fig:chess_example}
    \end{minipage}    
\end{figure}

\paragraph{\textbf{Chess puzzles}}

The third dataset we use is a corpus of chess puzzles. The data is furnished by Lichess, an online open-source chess server \citep{lichess}. From billions of games, Lichess compiles ``puzzles'' -- mid-game boards for which a sequence of unique best moves is determinable (e.g. sequences of moves leading to a forced checkmate). From this database, we compiled labeled data where the inputs are $8 \times 8 \times 12$ arrays indicating the position of each piece on the board (one channel per piece type and color) and the outputs are $8 \times 8$ binary masks showing the origin and destination positions for the optimal move. See the example in Figure \ref{fig:chess_example}, where it is white to move and the solution to the puzzle is to move the queen from F3 to F7.

The puzzles each have a difficulty rating determined by an Elo-like system (a standard system for rating players using tournament play) \citep{elo1978rating}. When Lichess users, each with an Elo rating, attempt to solve the puzzle, the rating of the player and puzzle is updated as if they ``played'' against each other. After enough players encounter the puzzle, the rating of the puzzle reaches an equilibrium which gets recorded in the database. We use these ratings to distinguish between easy and hard; 600,000 puzzles of difficulty rating less than 1,385 are used for training, and testing is performed on 100,000 examples with ratings greater than 1,385. 

Our models output a confidence score between zero and one for each square on the board. We take the two highest scores and compare their locations to the target to measure correctness -- only exact matches are considered correct. 

\section{Model architectures \& training}
\label{sec:archs}

In all of the experiments in this work, we employ network architectures based on ResNets \citep{he2016deep}. Our feed-forward networks are slight deviations from the most commonly used ResNets, in that the width does not change except at the first layer and after the last residual block, and we do not use batch normalization. This is done so that the recurrent models, whose internal recurrent module has the same input and output dimensions, can be as similar as possible to the feed-forward ResNets. In fact, the only difference during training between a feed-forward and recurrent model of the same effective depth is that the weights are shared between the residual blocks in the recurrent models. We refer to the recurrent portion of the network as the \emph{recurrent module}. Also, our models are fully convolutional with no fully connected heads. For solving prefix sums, which involves one-dimensional strings, we further deviate from classical ResNets by using one-dimensional convolutions.
For complete architectural details, see Appendix \ref{sec:app_architectures}.

We measure recurrent models in terms of \emph{iterations} and \emph{effective depth}. An iteration is a repetition of the recurrent residual block, which contains four layers in all of our models. Therefore, the effective depth is equal to four times the number of iterations, plus non-recurrent encoder and head layers that sandwich the recurrent module. For example, the models used for computing prefix sums have one convolutional encoder layer, followed by the recurrent block and then by a three layer convolutional head. In this case, a 10-iteration model has effective depth $1 + 10\times4 + 3=44$ layers.

For each training sample we consider, the label is an array of binary classification variables, one per input pixel. The training loss is simply the mean cross-entropy loss across output values. In general, hyperparameters were determined with the goal of finding convergent models. The specific batch sizes, learning rate and decay schedules, and other hyperparameter values are all available in Appendix \ref{sec:app_hyperparams}.\footnote{Code to reproduce our experiments along with information about downloading the data we use is available at \url{https://github.com/aks2203/easy-to-hard}.}

\section{Recurrent networks can generalize from easy to hard problems}
\label{sec:generalize}

We explore the ability of recurrent neural networks to generalize to more difficult problems simply by thinking deeper. To this end, we train models of varying effective depth on easy training examples and test them on harder problems. We find that recurrent models are even better at generalizing from easy to hard than their feed-forward counterparts. While there is only one way to test the feed-forward models, we take a closer look at what happens when the recurrent models are allowed to think deeper about the harder problems. Formally, we use more iterations of the recurrent module within the recurrent models when performing inference on test data. Across all three problem types, we find that the confidence of the model is a good surrogate for correctness. Therefore, when evaluating recurrent models, we use the output from the iteration to which the network assigns the highest confidence.

\subsection{Prefix sums}

The first task on which we demonstrate the ability of recurrent neural networks to learn an algorithm is one from the classical algorithms literature, namely computing prefix sums. Specifically, we study the problem of computing the prefix sums modulo two of binary input strings. 

When computing prefix sums, we employ models with effective depths from 40 to 68 layers. We train models on easy data consisting of 32-bit input strings and test on harder 40-bit and 44-bit strings. In Table \ref{tab:prefix_generalize}, it is clear that even when holding the depth constant at test time, recurrent models generalize from easy to hard better than feed-forward networks. 

\begin{table}[h]
    \centering
    \caption{\textbf{Extrapolating to longer input strings.} Shown here are the average accuracies of models trained on 32-bit inputs and tested on 40-bit inputs. The effective depths listed below correspond to 9, 10, and 11 iterations in recurrent models. We report average accuracy $\pm$ one standard error.}
    \label{tab:prefix_generalize}
    \small
    \begin{tabular}{lcccc}
    \toprule
         &  \multicolumn{3}{c}{Effective Depth (Layers)} \\
         & 40 & 44 & 48 \\
        \hline
        Recurrent & $24.96 \pm 2.96$ & $31.02 \pm 2.56$ & $35.22 \pm  3.34$ \\
        Feed-forward  & $22.17 \pm 0.85$ & $24.78 \pm 1.65$ & $22.79 \pm 1.32$  \\
        \bottomrule
    \end{tabular}
\end{table}

\begin{figure}[h!]
    \centering
    \includegraphics[width=0.8\textwidth]{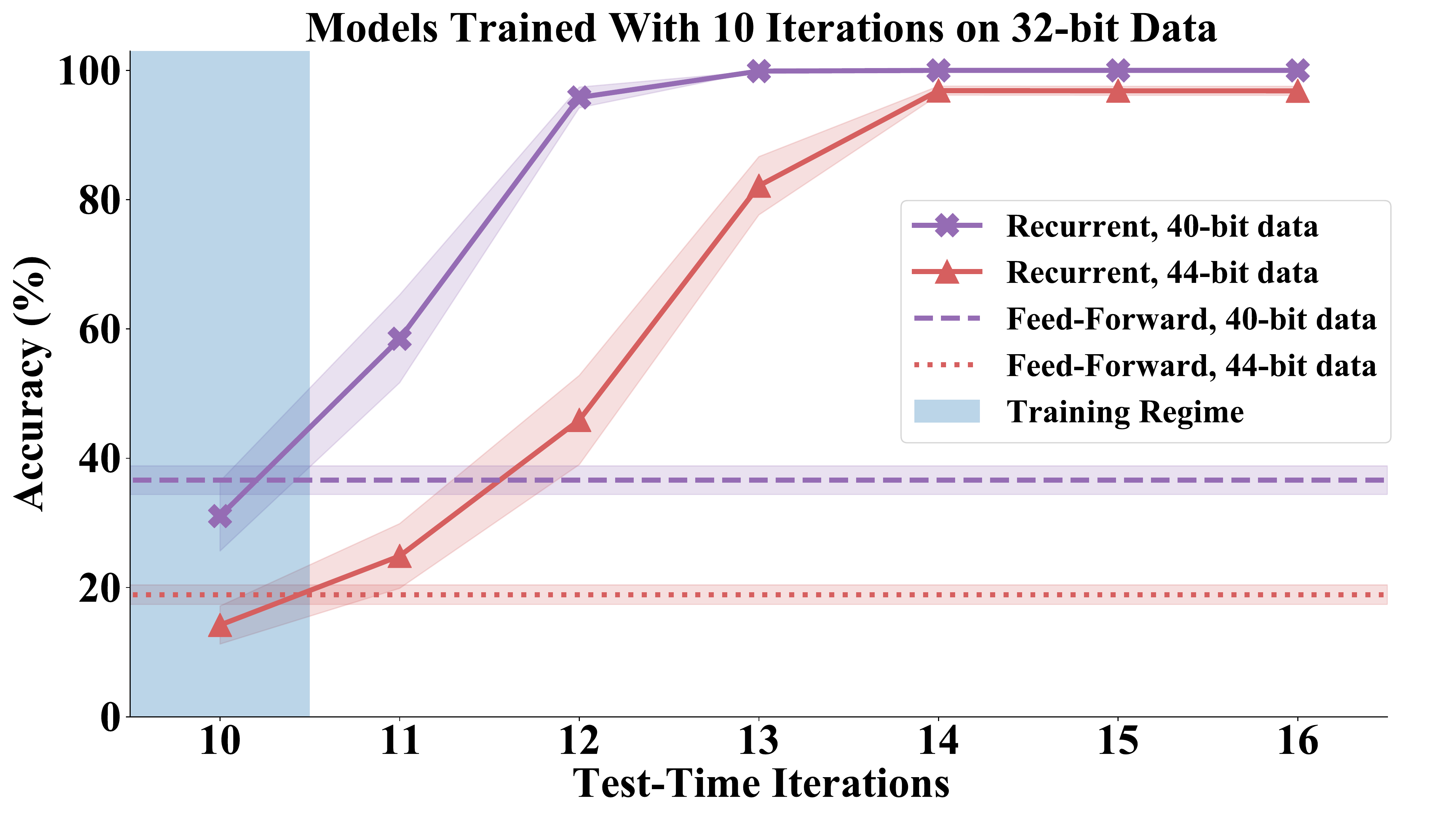}
    \caption{Generalizing from easy to hard prefix sums. The ability of networks to compute prefix sums on two test sets with longer input strings than were used for training (accuracy on 40-bit inputs in purple and on 44-bit inputs in red). We compare recurrent models to the best feed-forward models of comparable effective depth. The markers are at average values from several trials and the shaded regions indicate $\pm$ one standard error.}
    \label{fig:prefix_leap}
\end{figure}

When the thought budget, or number of iterations, is increased, we see that recurrent models can get upwards of 90\% of the harder testing examples correct. In Figure \ref{fig:prefix_leap}, we observe this large boost in the recurrent models' performance and a vast difference in the accuracy of recurrent models (with added iterations at test time) and feed-forward networks. Note that the dotted lines represent the average accuracy of the deepest feed-forward networks considered. That depth is 68 layers, or the effective depth of recurrent models with 16 iterations, and we use this baseline in the plot specifically because in the range of depths corresponding to the numbers of iterations shown, these feed-forward models achieve the highest accuracy. In other words, recurrent models trained with relatively few iterations generalize well to harder data while similar and even much larger feed-forward networks fail to generalize in the same scenario.

The generalization we see in Figure \ref{fig:prefix_leap} indicates that these recurrent models learn processes that can be extended to harder problems by running for more iterations. In particular, the recurrence is both the machinery that allows for varying the depth at test time, as well as a force at training to push the model to find parameters that make progress toward a solution with each reuse. 

When these results are viewed through the lens of algorithm design, one might wonder how the \emph{receptive field}, or the number of entries in the input that determine a single entry in the output, affects these models.
The feed-forward models, whose accuracies are shown in Figure \ref{fig:prefix_leap}, have the same receptive field as the recurrent models when tested with 16 iterations. This makes it clear that the increase in accuracy of recurrent models does not simply occur because the receptive field grows with added iterations, rather it occurs because they have learned a process that can extrapolate beyond the training distribution. Further discussion on receptive field is presented in Appendix \ref{sec:app_further}.

\subsubsection{Iterative outputs}

\begin{figure}[h!]
    \centering
    \includegraphics[width=0.7\textwidth]{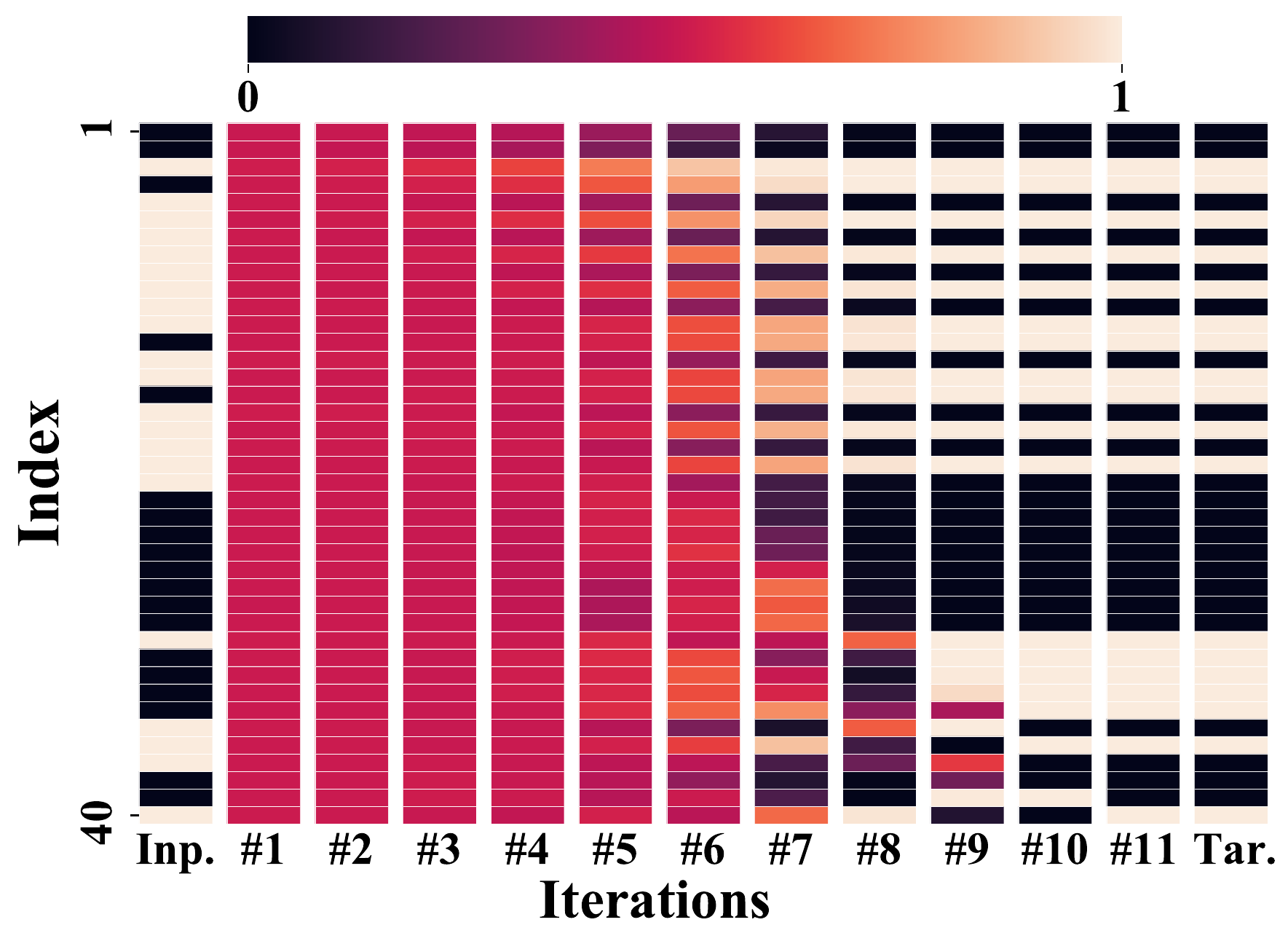}
    \caption{A recurrent model's output from each of 11 iterations on a 40-bit input string. Shown here is the confidence that there is a 1 at each index of the output. The first index is at the top for all vectors, the input is in the left-most column and the  target is in the right-most column. The model used to produce this plot was trained with fewer iterations (10) on shorter input strings (32-bit).}
    \label{fig:prefix_thoughts}
\end{figure}

One way to dissect the learned process and compare it to known algorithms is to plot the confidence of the model at each iteration. In Figure \ref{fig:prefix_thoughts}, we show a representative example of a network's confidence that each bit in the output is a one. Two striking observations can be made from this figure. The first is that the model is progressing to the solution with each iteration. The second observation is that it is resolving the prefix sum in the earlier bits first and moving down the string, settling on the final bits only in the last iteration. This is remarkably similar to a naive algorithm one might implement for this task that marches from the first index until the end of the string computing prefix sums in order.

\subsection{Mazes}

For maze solving, we train models on a training set composed of the easier small mazes, and we investigate the ability of networks to make the leap to larger, or harder, mazes at test time.  In line with the findings above, we show two important behaviors. First, the recurrent models make the leap from small to large mazes better than feed-forward models. Second, when allowed to think deeper, the recurrent models exhibit even higher performance. 

The networks employed here are fully convolutional and have 512 channels in the internal layers. In assessing the confidence of a given output, we average each pixel's classification confidence.

\begin{table}[h!]
    \centering
    \caption{\textbf{The average accuracy (\%) of models trained on small mazes and tested on large ones.} Over a range of effective depths, we see that recurrent models generalize to the harder mazes better than their feed-forward counterparts. Figures reflect averages over several trials $\pm$ one standard error.}
    \label{tab:mazes_generalize}
    \small
    \begin{tabular}{rccccc}
    \toprule
         &  \multicolumn{5}{c}{Effective Depth} \\
         & 20 & 24 & 28 & 36 & 44 \\
        \hline
        Recurrent & $12.66 \pm 0.44$ &   $14.02 \pm 0.39$ & $19.95 \pm 0.31$ & $22.96 \pm 1.03$ &   $29.72 \pm 1.22$ \\
        Feed-forward  & $7.94 \pm 0.36$ &  $12.43 \pm 0.50$ & $14.67 \pm 0.54$ & $17.71 \pm 0.36$ &   $22.53 \pm 1.14$ \\
        \bottomrule
    \end{tabular}
\end{table}

\begin{figure}[h!]
    \centering
    \includegraphics[width=0.8\textwidth]{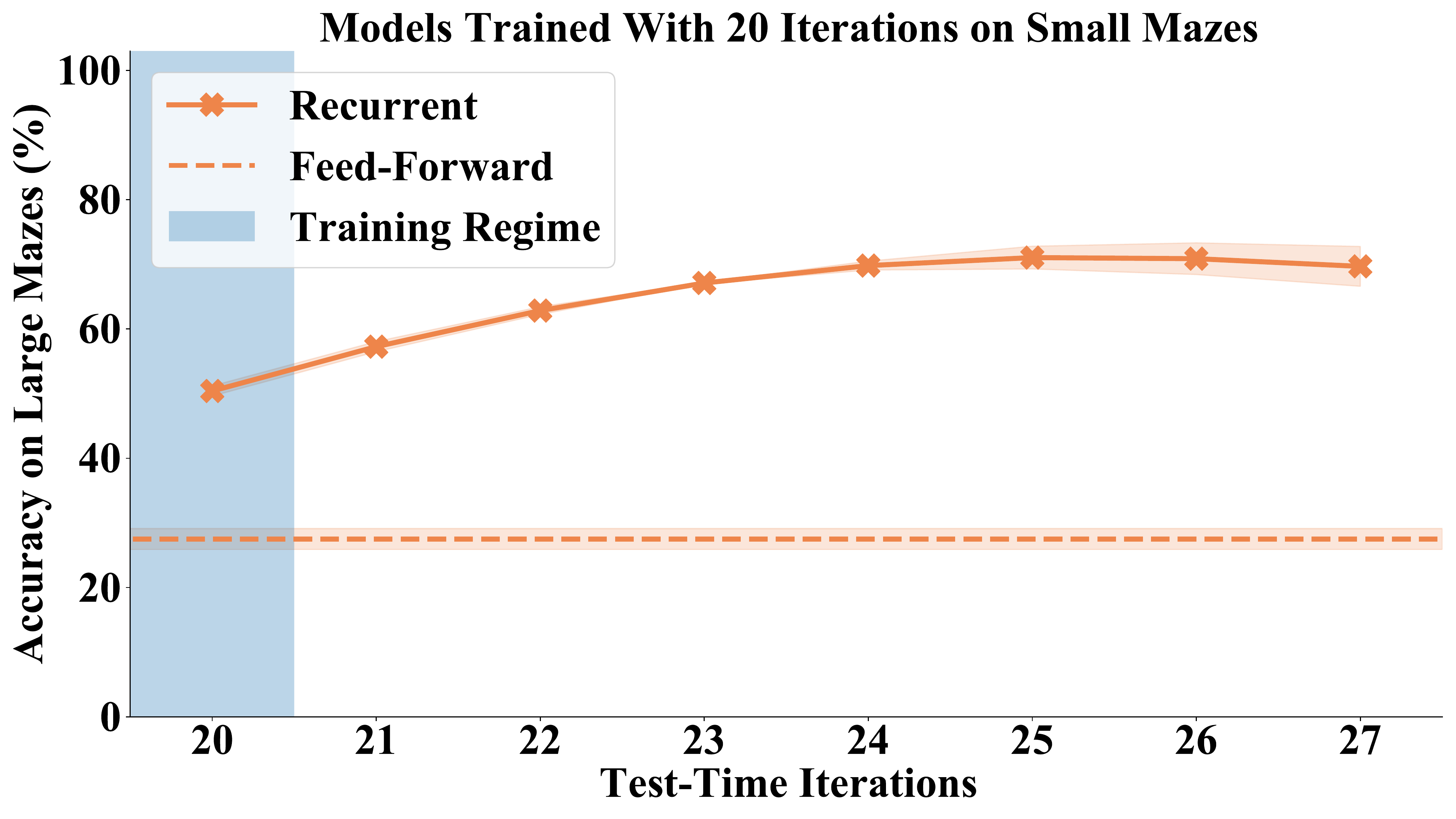}
    \caption{Generalizing from easy to hard mazes. We compare recurrent models to the best feed-forward models. The markers are at average values from several trials and the shaded regions indicate $\pm$ one standard error.}
    \label{fig:mazes_leap}
\end{figure}

Table \ref{tab:mazes_generalize} shows that for a fixed effective depth, recurrent models always generalize to the hard mazes better than their feed-forward counterparts.
Inspired by the upward trend in Table \ref{tab:mazes_generalize}, we shift focus to deeper models. In Figure \ref{fig:mazes_leap}, we show that recurrent models can extrapolate to harder problems better than feed forward models. When trained on small mazes with 20 iterations (effective depth of 84 layers), these networks can solve about half of large mazes. However, when allowed to think for longer, the recurrent models can correctly solve an even higher proportion of large mazes. In fact, models trained with 20 iterations can achieve upward of 70\% accuracy on large mazes using 5 additional iterations at test time. 

Maze solving is another task for which global information is needed. We investigate how dilated filters affect model performance. Dilation is a way of changing the receptive field without adding new parameters or more depth. Table \ref{tab:mazes_dilated} shows that dilations lead to slight improvements, however, the benefits of recurrence are still abundantly clear. Indeed, the difference in performance is even larger. 

\begin{table}[h!]
    \centering
    \caption{\textbf{The average accuracy (\%) of models with dilated filters trained on small mazes and tested on large ones.} Figures reflect averages over several trials $\pm$ one standard deviation.}
    \label{tab:mazes_dilated}
    \small
    \begin{tabular}{rccccccc}
    \toprule
         &  \multicolumn{5}{c}{Effective Depth} \\
         & 20 & 24 & 28 & 32 & 36\\
        \hline 
        Recurrent & $33.60 \pm 1.06$ & $40.49 \pm 1.63$ & $33.91 \pm 1.99$ & $40.56 \pm 4.34$ & $50.50 \pm 7.97$ \\
        Feed-forward & $19.73 \pm 0.72$ & $21.59 \pm 0.22$ & $24.18 \pm 1.33$ & $25.82 \pm 0.10$ & $26.54 \pm 2.13$ \\
        \bottomrule
    \end{tabular}
\end{table}


\subsubsection{Iterative outputs}

\begin{figure}[h!]
    \centering
    \raisebox{0.09\height}{\includegraphics[width=0.24\textwidth]{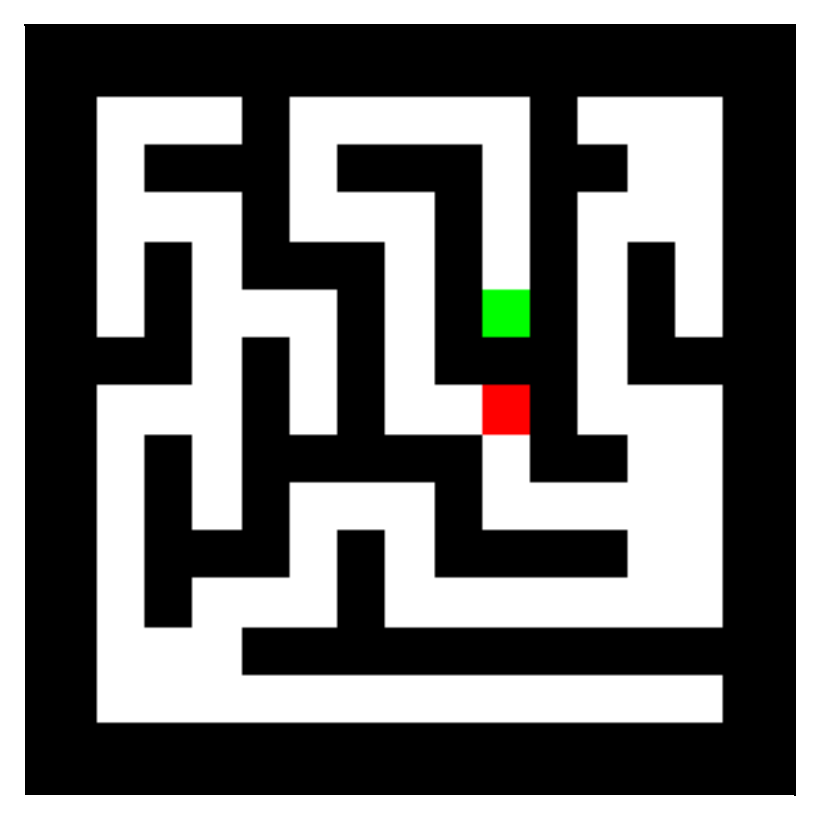}}
    \includegraphics[width=0.66\textwidth, trim=0cm 0.25cm 0cm 0cm, clip]{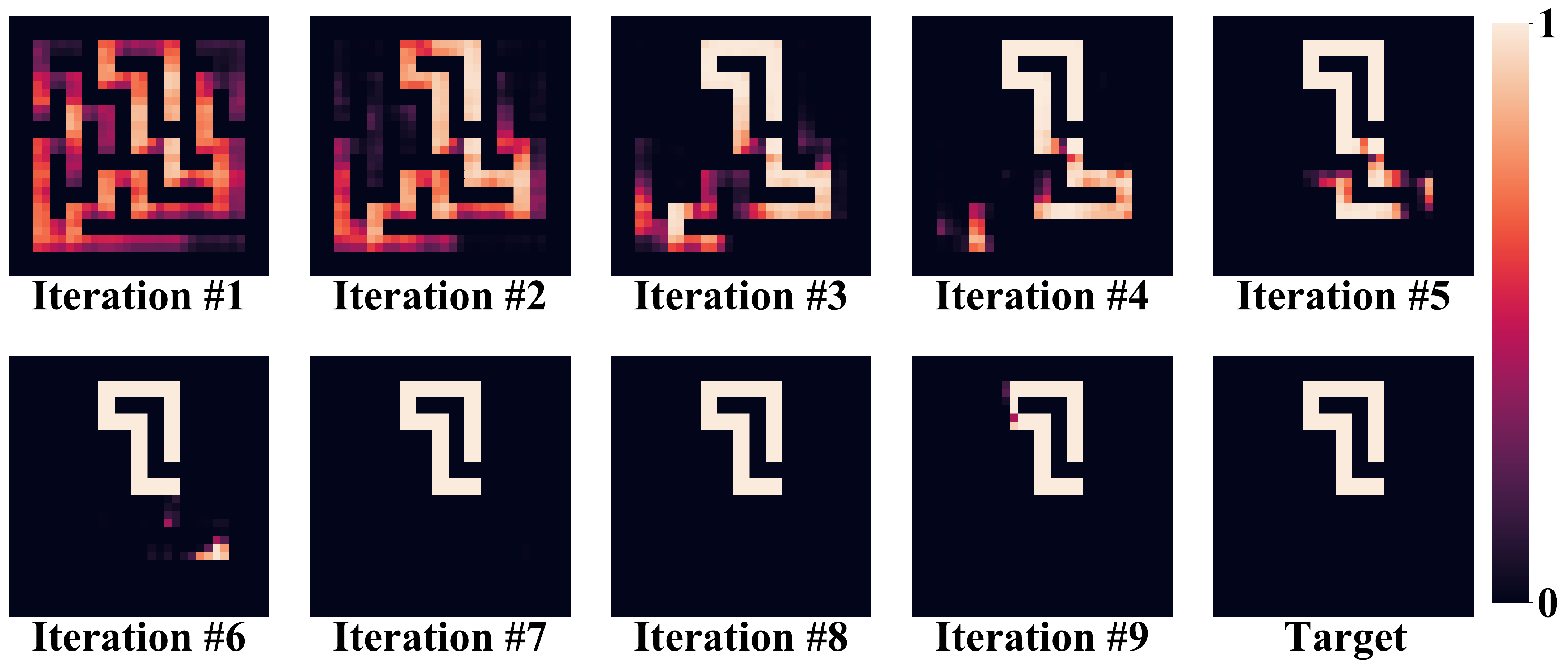}
    \caption{Input, target, and outputs from different iterations are shown to highlight the model's ability to think sequentially about mazes. We plot the model's confidence that each pixel belongs to the optimal path. This is a representative example from a model trained to solve small mazes in six iterations.}
    \label{fig:mazes_thoughts}
\end{figure}

The recurrent maze solving networks also produce output at every iteration. Examining this output again leads to a remarkable conclusion: these recurrent models are narrowing in on the answer with each successive iteration. In Figure \ref{fig:mazes_thoughts}, it is clear from the output on iteration four that the network has found two routes emanating from the red square. Moving through the iterations, the model refines the output, increasing the confidence for pixels on the path and decreasing the others until finally at Iteration \#7, the output matches the target. It is also interesting to observe here that the output is consistently correct for two iterations, after which a few pixels flip (Iteration \#9).

\subsection{Chess puzzles}

The third dataset comprises chess puzzles, or mid-game chess boards for which we seek the best next move. Unlike the other two datasets, the state of the art for chess playing algorithms is complex and has components that use algorithms like Monte Carlo tree search as well as neural network based elements for evaluating positions \citep{stockfish, silver2017mastering}. The complicated nature of these systems provides some context for how difficult these puzzles are. What makes these puzzles particularly useful for us is that there is a predetermined best next move. A move is defined as an origin square, or the current location of the piece to be moved, and a destination square.\footnote{There are some cases, pawn promotions, where this information does not uniquely identify the move, and they are overlooked in this project.} In order to generate target outputs for our models, we define a move as an $8\times 8$ array with zeros everywhere except at the entries corresponding to the origin and destination squares which are ones.

We compare recurrent and feed-forward networks of effective depths from 84 layers to 100 layers that take $8 \times 8 \times 12$ arrays as input. These fully convolutional networks have 512 channels in the internal layers, and the output is $8\times 8 \times 2$, corresponding to binary classification at each input pixel. During training, we use an average of cross-entropy losses at every pixel. When evaluating these models, however, we define the predicted move by the locations of the two highest confidence scores.

\begin{figure}[h!]
    \centering
    \includegraphics[width=0.8\textwidth]{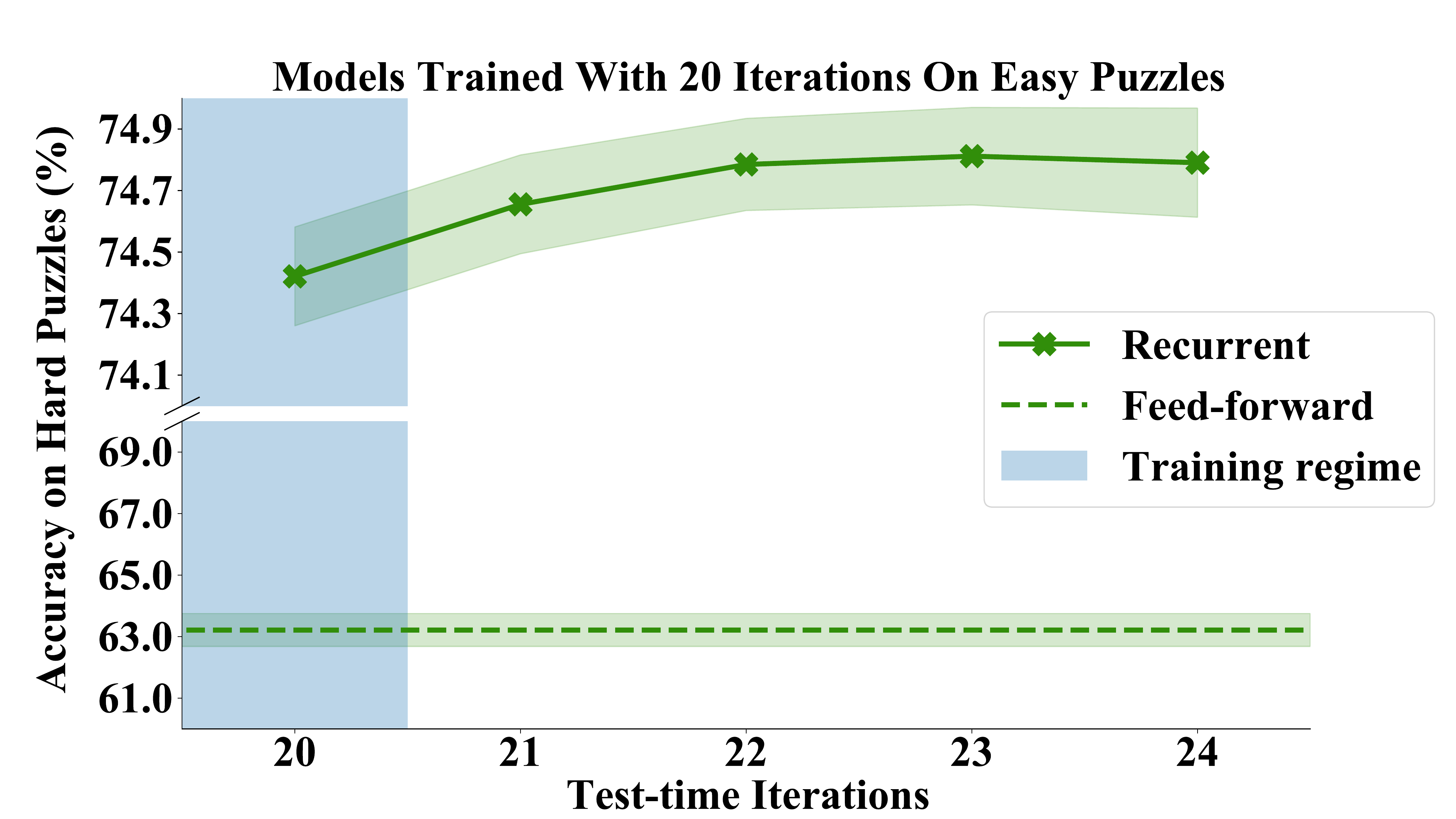}
    \caption{Generalizing from easy to hard chess puzzles. The ability of networks to solve harder puzzles than were used for training. We compare recurrent models to the best feed-forward models of comparable effective depth. The markers are at average values from several trials and the shaded region indicate $\pm$ one standard error.}
    \label{fig:chess_leap}
\end{figure}

Once again, we see that recurrent models can solve more chess puzzles than their feed forward counterparts. Furthermore, by thinking deeper at test time, recurrent models can perform even better. While the gains shown in Figure \ref{fig:chess_leap} are modest in comparison to the other two problem settings, the trend is clear -- recurrent models can solve more puzzles with more iterations.

\subsubsection{Iterative outputs}

\begin{figure}[h!]
    \centering
    \raisebox{0.1\height}{\includegraphics[width=0.26\textwidth]{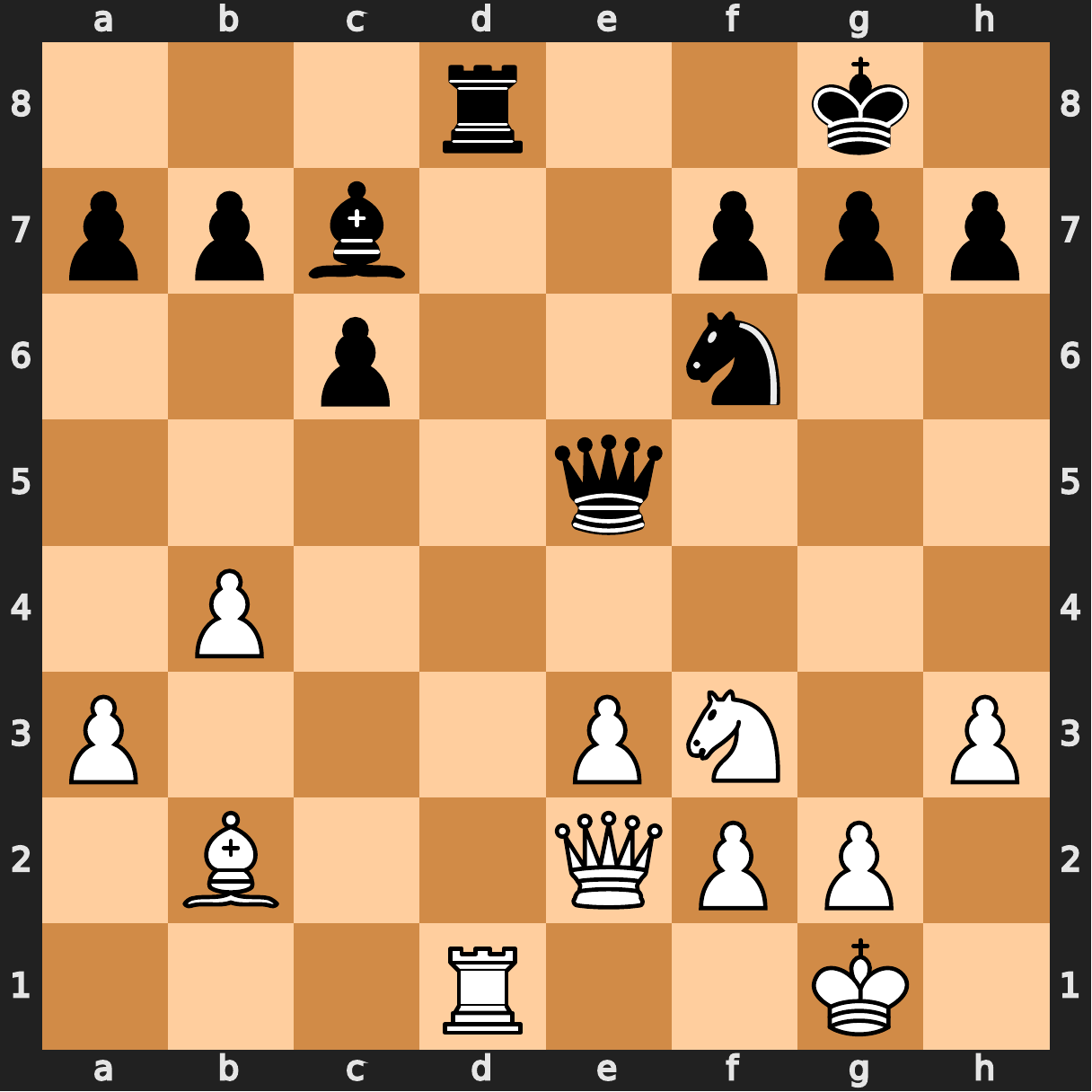}}
    \includegraphics[width=0.64\textwidth, trim=0cm 0.25cm 0cm 0cm, clip]{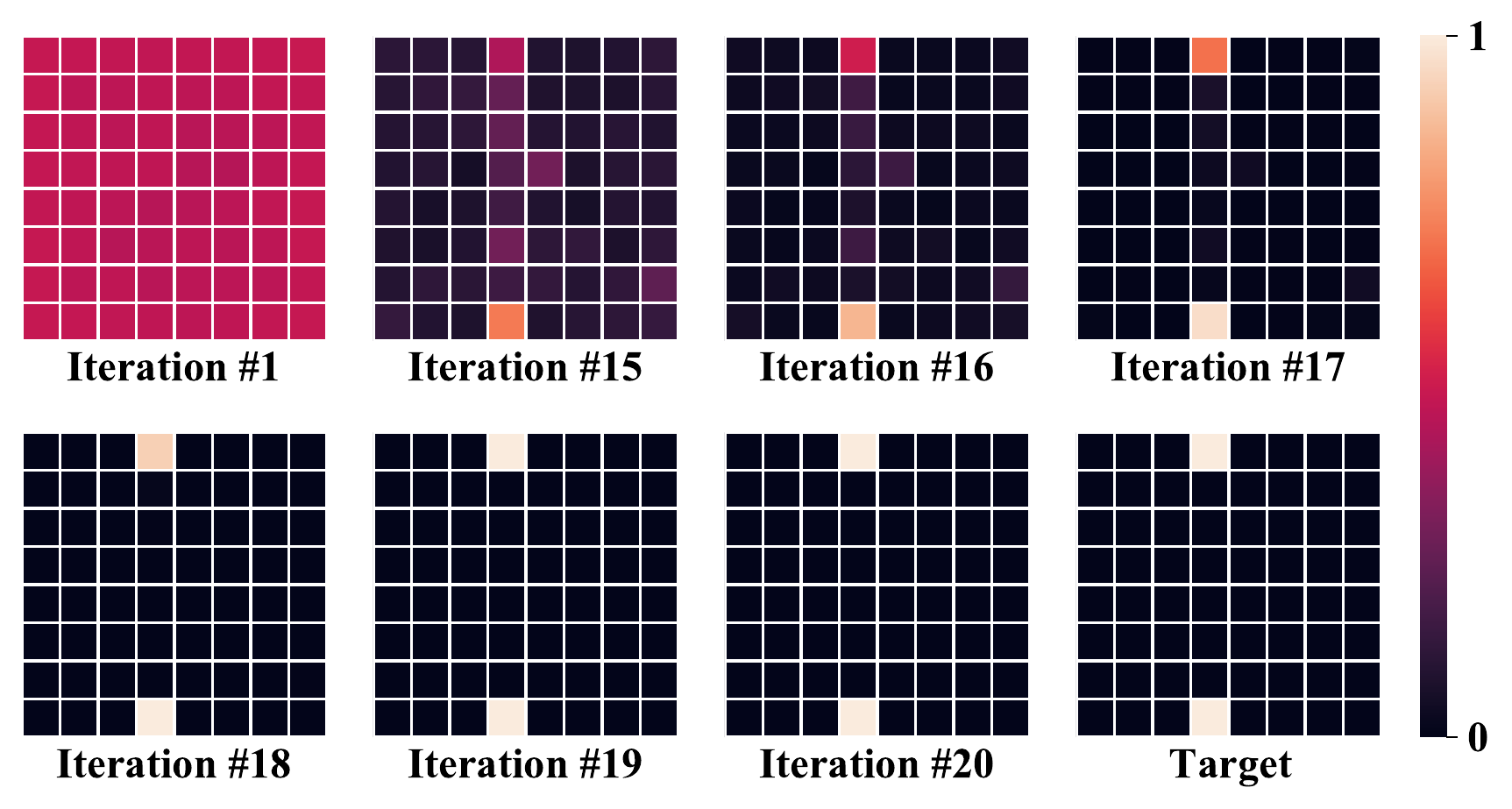}
    \caption{Input, target, and outputs form different iterations are shown to highlight the model's ability to think about the next move. We plot the model's confidence that each pixel is one of the two that define a move. In this example, black is to move next. For space consideration, iterations 2-14, which look like the first iteration, are left out of this plot. More examples are available in Appendix \ref{sec:app_visualizations}.}
    \label{fig:chess_thoughts}
\end{figure}

While the outputs from the other two datasets show the similarity between the learned iterative process and known sum and search algorithms, extracting that insight on chess puzzles is much more difficult. Partly, this is because exploiting known search algorithms on such a large search space is hard to visualize. Nonetheless, the network's output, shown in Figure \ref{fig:chess_thoughts}, tells a fascinating story. First, the Iteration \#1 plot shows that after one iteration, we observe equal confidence at every location on the board. In the next frame, Iteration \#15, it is clear that the model is considering moving the D8 rook to multiple squares, and it also considers the intuitive idea of using the E5 queen to place the white king in check on H2. With each successive iteration, the network becomes more confident -- first that the rook is the correct piece to move, and then where to move it to.

\section{Discussion}
\label{sec:discussion}

More than an answer, the results and conclusions in this paper are posing a question: Can learned models behave like classical algorithms in the way they generalize to harder or larger problems?

Our discussion of this question and our observations begins with delineating the limitations of our work. The first major limitation is that we do not propose a definitive answer. Rather,
using representative cases, we demonstrate that recurrence can help neural models make the leap from easy training data to hard testing examples. A more subtle limitation of our work lies in how we split the data by difficulty. For prefix sum computation and for mazes, the classical algorithms approach to measuring problem complexity is tied to problem size, so in those settings we make intuitive easy/hard splits. With chess however, the issue is much more complex. Should puzzles with higher ratings require more memory or more computation? We carry out our work assuming the answer is yes, but we remain open minded to the possibility that the ratings assigned by Lichess may be weak surrogates for algorithmic complexity of each puzzle. In short, chess is an extremely difficult domain to analyze.

The observations above indicate that iterative models can learn processes that generalize beyond the training distribution with more iterations. One exciting use case is when the testing distribution is inaccessible, a setting where training on the harder distribution itself would be impossible. Many real life scenarios demand exactly this type of problem solving, from robots deployed in the real world after training in simulation to humans who spend a lot of time practicing on easy math problems only to spend years on difficult unsolved questions. 

On a conceptual level, the recurrent model behavior we show is analogous to the human behavior of manipulating representations in working memory; in this analogy,  the recurrent block performs the transformations and the activations it generates are the memory.  It is not the goal of the experiments here to suggest that iterative models use mechanisms similar to those in a human brain. Nonetheless, it is exciting to see, even in a proof of concept setting, neural models that appear to deliberate on a problem until it is solved and can extend their abilities by thinking for longer.

This raises a questions that motivates future work. Can we build neural networks that can think for even longer? Is it feasible to have models whose performance only increases with added compute time? Humans who are given more than enough time to solve a maze, will not suddenly get it wrong after arriving at the right answer, perhaps this awareness of when to stop thinking can be built into networks like the ones we study here.

\section{Conclusion}
In this work, we demonstrate that neural networks are capable of solving sequential reasoning tasks and then extrapolating this knowledge to solve problems of greater complexity than they were trained on.  These recurrent models are largely inspired by the classical theory of mind, in which the brain iteratively applies primitive strategies to solve complex problems over time \citep{baddeley2012working}. Our models are recurrent versions of popular architectures and we acknowledge that variations to the model may be helpful. Thus, we leave an in depth investigation into other neural network designs for future work.

Interestingly, the resulting models excel at solving problems that are classically solved by hand-crafted algorithms; prefix sums are computed using reduction trees, mazes are classically solved by depth/breadth first search, and chess is solved by Monte-Carlo tree search.  Even with the advances in machine learning that we have today,  hand-crafted algorithms still play a role in state-of-the-art reasoning systems.  A prominent example is AlphaZero, which plays board games using Monte-Carlo tree search algorithms assisted by a learned pruning function \citep{silver2017mastering}.  While moving away from this paradigm that includes hand-crafted elements is highly ambitious, this work suggests that it may be possible to train gameplay systems without building them on top of a hand-crafted tree search engine.  In other words, it may be possible to machine-learn these algorithmic behaviors end-to-end.

\section*{Acknowledgements}
This project was supported by the ONR MURI program, AFOSR MURI program, the DARPA Young Faculty Award, and the National Science Foundation Division of Mathematical Sciences.  Additional support was provided by Capital One Bank and JP Morgan Chase. Additionally, we thank Avrim Blum for thought-provoking and insightful conversations.

\bibliographystyle{plainnat}
\bibliography{main}

\clearpage
\appendix
\input{app}

\end{document}

%% file: app.tex
\section{Technical details}

\subsection{Datasets}
\label{sec:app_datasets}

Details of the datasets we introduce are presented in this section. Specific details about generation as well as statistics from the resulting datasets are delineated for each one below. The datasets can all be downloaded from \url{https://github.com/aks2203/easy-to-hard-data} \citep{schwarzschild2021datasets}.

\subsubsection{Prefix sum data}
\label{sec:app_prefix_data}

Binary string inputs of length $n$ are generated by selecting a random integer in $[0, 2^n)$ and expressing its binary representation with $n$ digits. Datasets are produced by repeating this random process 10,000 times without replacement. Because the number of possible points increases exponentially as a function of $n$ and the size of the generated dataset is fixed, it is important to note that the dataset becomes sparser in its ambient hypercube as $n$ increases. Moreover, we are limited to datasets with binary strings of length $n > 13$ to avoid duplicate data points.

\subsubsection{Maze data}
\label{sec:app_maze_data}
 
The maze data is generated using a depth first search algorithm. A grid is initialized with walls at every cell boundary. Then, using depth first search from a starting cell, every cell is visited at least once, removing walls along the path. The algorithm is available in the attached code. The resulting dataset has non-uniformly distributed path lengths. Also, this process does lead to duplicates, but fewer than $0.5\%$ of points are duplicated and so this is ignored in our study.

\begin{figure}[h!]
\centering
    \includegraphics[width=0.23\textwidth]{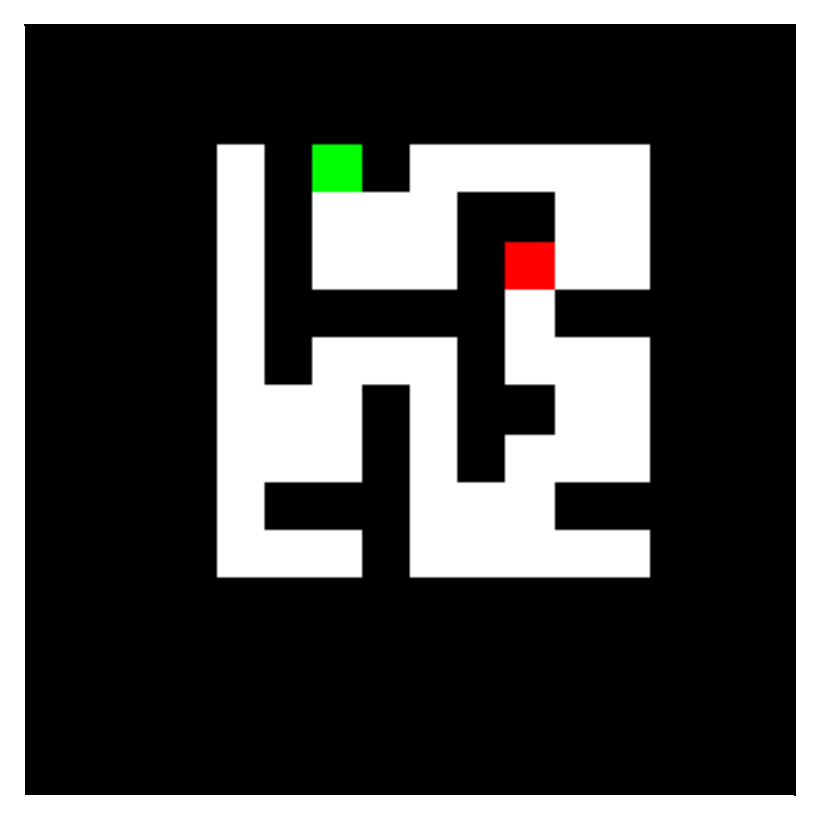}
    \includegraphics[width=0.238\textwidth, trim=1cm 1.1cm 1cm 1cm, clip]{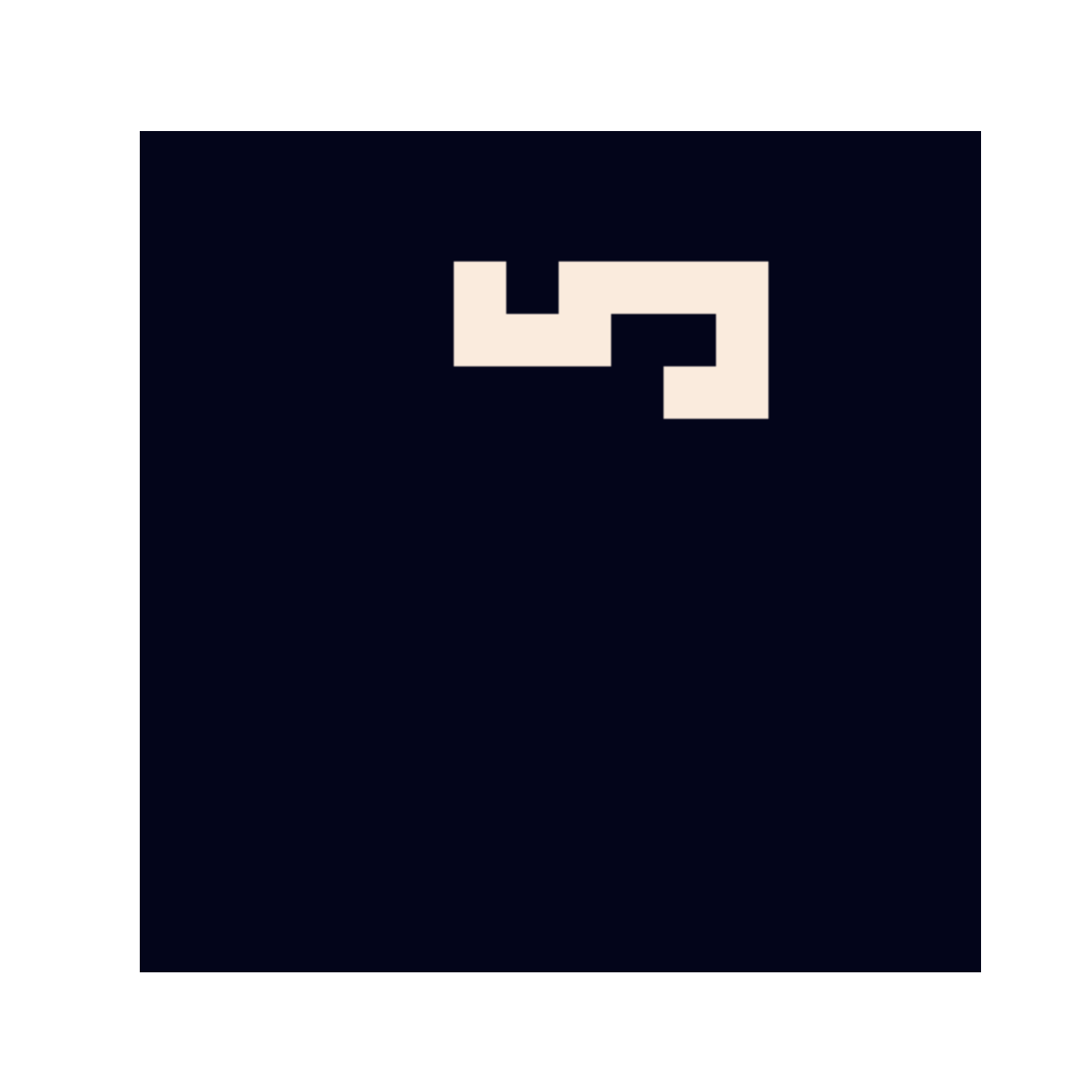}
    \includegraphics[width=0.23\textwidth]{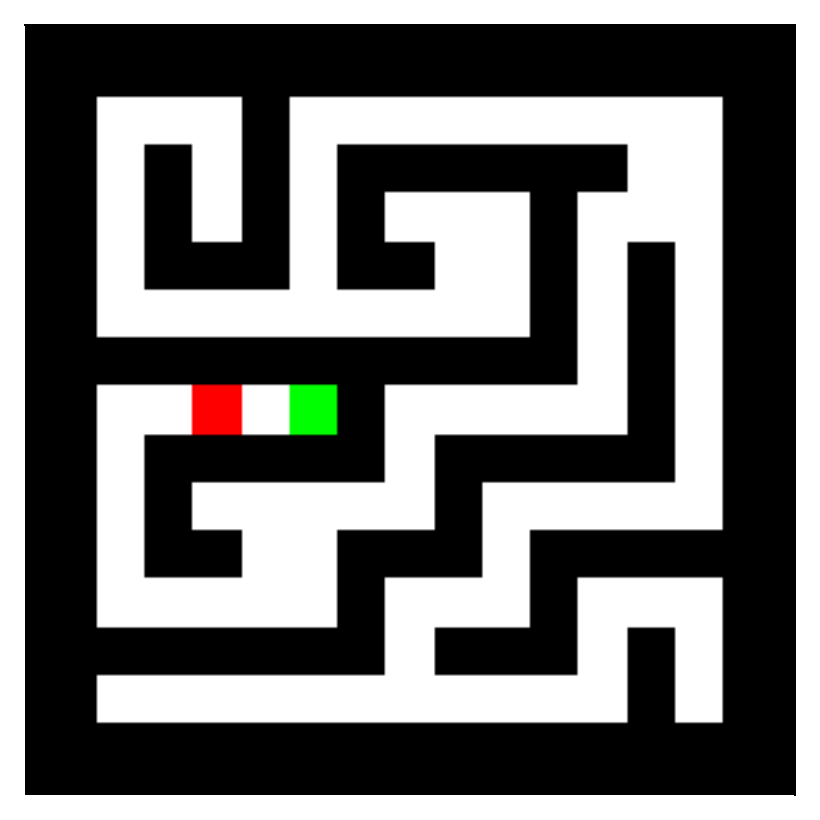}
    \includegraphics[width=0.238\textwidth, trim=1cm 1.1cm 1cm 1cm, clip]{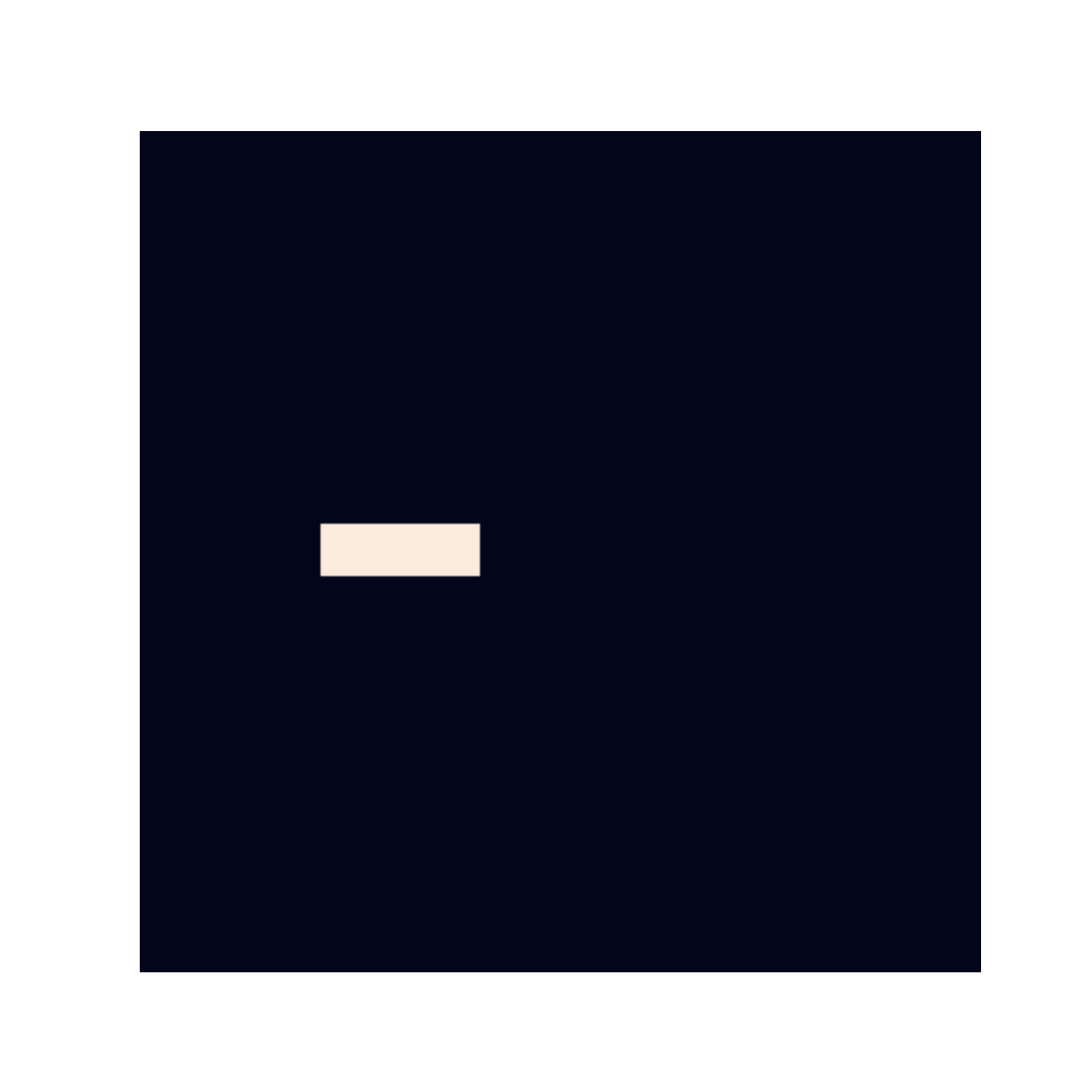}
    \caption{Example of small (left) and large(right) maze inputs and targets. The target is a binary classification label for each pixel indicating on/off the optimal path.}
    \label{fig:maze_example}
\end{figure}

\subsubsection{Chess puzzle data}
\label{sec:app_chess_data}

Starting position data from the Lichess database in FEN form are transformed into $8 \times 8 \times 12$ arrays indicating the position of each piece on the board (one channel per piece type and color) to be used as inputs. The first correct output moves in UCI move format are transformed into $8 \times 8$ binary masks showing the origin and destination positions for the optimal move and used as targets for the corresponding input. More specific details about these transformations can be viewed the provided code.

Lichess compiles puzzle data by analyzing 200,000,000 from their database of games between users using the Stockfish 12/13 NNUE chess engine at 40 meganodes. Though amalgamated puzzles do not come with username information, this information can ostensibly be gathered by playing through the puzzles on the Lichess website, where this information \emph{is} available. Because of this, applying deep learning to this dataset could reveal some information about the playing style of Lichess users. However, upon creating accounts users agree to allow their game data to be public and be used specifically for investigative purposes.

The puzzle data is released by Lichess for public use under the Creative Commons CC0 license. 

\subsection{Architectural details}
\label{sec:app_architectures}

Model files are available in the linked code repository, their details are as follows.

The feed-forward prefix sum models are fully convolutional models that take in $n \times 1$ arrays. The first layer is a one-dimensional convolution with a three entry wide kernel that strides by one entry with padding by one on either end on the input. The output of this first convolution has 120 channels of the same shape as the input. The next parts of the networks are residual blocks made up of four layers that are identical to the first layer with skip connections every two layers. After the residual blocks, there are three similar convolutional layers that output 60, 30, and two channels, respectively. For a network of depth $d$, there are $(d-4)/ 4$ residual blocks. The recurrent models are identical, except that all residual blocks share weights.

The feed-forward maze solving models are fully convolutional models that take in $n \times n \times 3$ arrays. The first layer is a two-dimensional convolution with a $3 \times 3$ kernel that strides by one entry and pads by one unit in each direction. The output of this first convolution has 128 channels of the same shape as the input. As above, the next parts of the networks are residual blocks made up of four layers that are identical to the first layer with skip connections every two layers. After the residual blocks, there are three similar convolutional layers that output 32, 8, and two channels, respectively. For a network of depth $d$, there are $(d-4)/ 4$ residual blocks. The recurrent models are identical, except that all residual blocks share weights.

For experiments with dilated filters, the only changes made are to the dilation of the convolutional filters and to the padding of every convolution and the values are set to maintain the output dimenssion of each layer.

The chess playing models are the same as the maze models except that the first layer takes $8 \times 8 \times 12$ inputs and outputs 512 channels.

None of the models used in this project have batch normalization or bias terms.

\subsection{Training hyperparameters}
\label{sec:app_hyperparams}
The training details and hyperparameters are outlined below.
\begin{itemize}
    \item Prefix sum training is unstable, thus we only save models that show 100\% training accuracy at the end of training.
    \item Data augmentation: Binary string inputs to prefix sum networks are approximately normalized by subtracting 0.5 from every element in the string in order to aid in training stability. Mazes are padded to be $32 \times 32$ pixels. 
    \item Optimizer: All prefix sum networks are trained using the Adam optimizer with a weight decay factor of 2e-4. Because of training instability, we also apply gradient clipping at magnitude 1.0. The maze solving networks are trained with stochastic gradient descent with a weight decay factor of 2e-4 and momentum coefficient of 0.9. Chess networks are trained using stochastic gradient descent with a weight decay factor of 2e-4 and momentum coefficient of 0.9.
    \item Epochs: Prefix sum networks are trained to convergence with 500 epochs. Maze models are trained for 200 epochs. Chess networks are all trained to convergence with 140 epochs.
    \item Learning rate and decay schedule and type. All prefix sum networks are trained using an exponential warm-up schedule applied over 10 epochs. Initial learning rate (post-warmup) is set at 0.001 and is subsequently halved at epochs 100, 200, and 300. Maze solving networks also use warm-up with a period of 5 epochs after which the learning rate is 0.001. The learning rate further decays by a factor of ten at epoch 175. Chess networks are trained with an exponential warm-up schedule applied over 3 epochs. Initial learning rate (post-warmup) is set to 0.1 and dropped by a factor of ten at epochs 100 and 110.
    \item Batch size: For training prefix sum models, we use batches of 150 binary strings. When training maze networks we use batches of 50 mazes. For chess models, we train with batches of 300 puzzles.
\end{itemize}

\subsection{Compute resources}
\label{sec:app_compute}

All of our experiments were done on Nvidia GeForce RTX 2080Ti GPUs. Prefix sum models train in less than one hour on a single GPU, whereas maze solving models require approximately seven hours on a single GPU. The chess networks were trained in about 24 hours on four GPUs.

In total, the prefix sum training required to generate the results presented in this paper takes approximately 30 GPU hours. The maze models, in total, require one GPU week. And the chess networks took 3 GPU weeks.

The data pre-processing as well as the testing of all models for every experiment can done comparatively quickly, in several hours.

\section{Further insights}
\label{sec:app_further}

We present several additional findings from our experiments. 

\subsection{Deeper feed-forward models}

Significantly deeper feed forward models cannot generalize from easy to hard as well as recurrent ones. For example, we extend the experiments in Figure 4 to feed-forward models with depths 132, 264, and 528. These models all fit the training data (32-bits) well, but the best performance on 40-bit data is 53\% and on 44-bit data is 27\% -- neither is attained by the deepest models lending confidence that the performance would not increase with even more depth. 

\subsection{In-distribution tests}

For in-distribution test accuracy (on small/easy cases) on prefix sums, both classes of models achieve >99\% and on mazes, both achieve >97\% (using the same number of iterations as used during training). When we apply more iterations, we see only slight drops for in-distribution test accuracy (tenths of a percent), but no improvement. On chess puzzles, the in-distribution test accuracies are 92\% for the recurrent models and 84\% for the feed-forward networks (though there is a gap in performance, it is smaller than the gap observed for out-of-distribution testing).

We also train and test on the harder datasets to measure in-distribution accuracy. Here, we find there is little difference between recurrent and feed-forward models. As examples, both classes of models for prefix sums and mazes achieve essentially 100\% on unseen data. Specifically, both recurrent and feed-forward models achieve >99\% on prefix sums and >97\% on mazes. 

Finally, we train models on larger instances and test them on smaller cases, and we find that they can solve smaller examples in fewer iterations than were used at training. For example, when models are trained to compute prefix sums on 44-bit data in 10 iterations, they can get 100\% of the 16-bit test examples correct after only 4 iterations.

\subsection{Prefix sum experiments on other datasets}
\label{sec:app_prefix_extra_exps}

\begin{figure}[h!]
    \centering
    \includegraphics[width=0.8\textwidth]{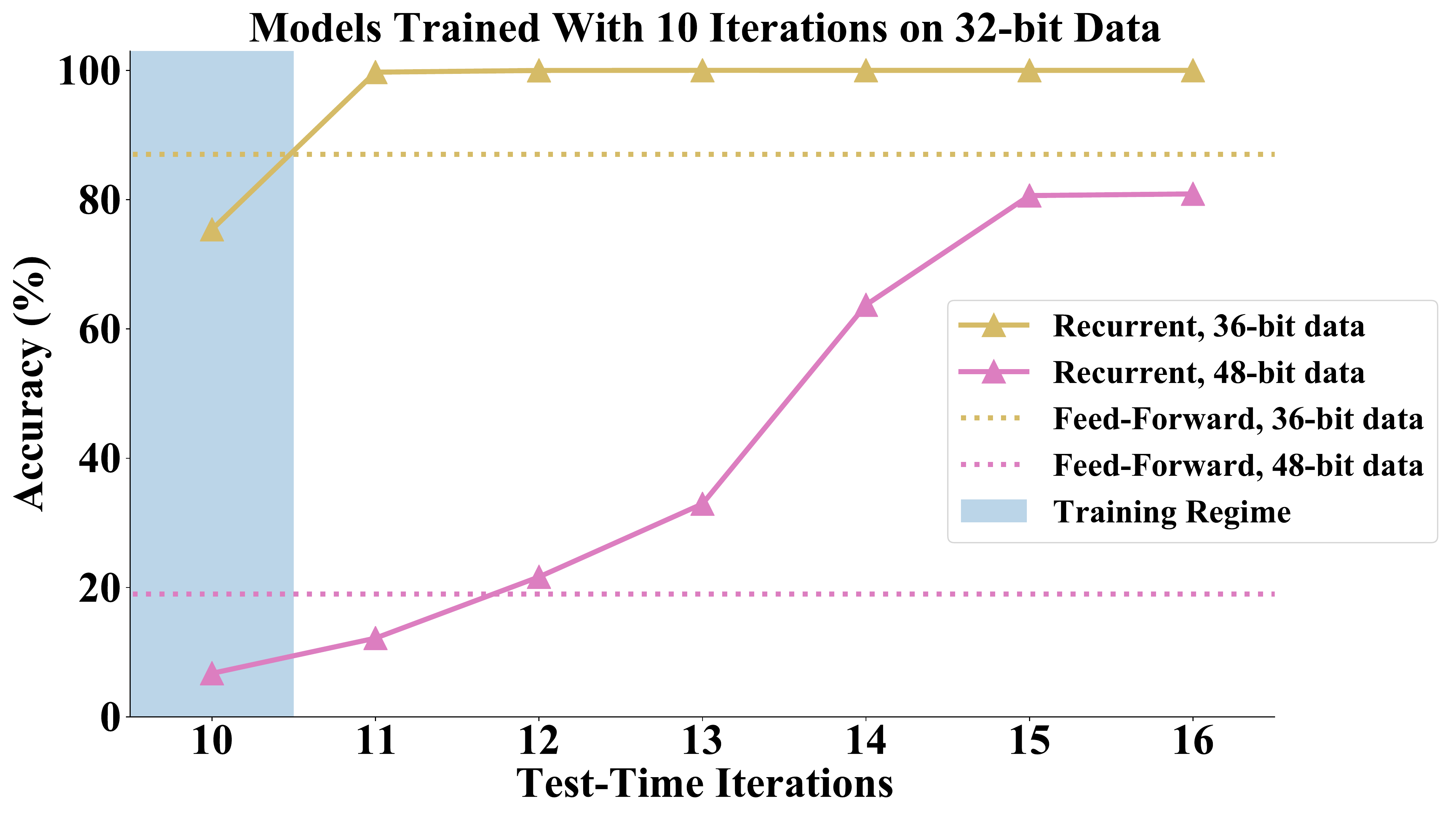}
    \caption{Generalizing from easy to hard prefix sums. The ability of networks to compute prefix sums on two additional test sets with longer input strings than were used for training (accuracy on 36-bit inputs in yellow and on 48-bit inputs in pink). We compare recurrent models to the best feed-forward models of comparable effective depth.}
    \label{fig:app_prefix_leap}
\end{figure}

\subsection{Dilated filters}
\label{sec:app_dilated}

The receptive field can be increased without adding parameters or depth by dilating the convolutional filters. When we use dilated filters to compute prefix sums, we find that we can fit the training data with $N$-bit sequences with fewer than $N$ layers -- a behaviour that is not possible with non-dilated convolutions. This suggests that the algorithm learned by our models is informed by the receptive field. In other words, since the final entry in a prefix sum does require global information (and therefore, in order for a neural network to compute the final entry it needs a complete receptive field), the use of extra iterations on longer sequences meshes with the algorithmic analysis. When we contextualize our models by analyzing the way they scale, it is reasonable to find that networks with non-dilated convolutions scale linearly with the input length. This also motivates future work to study whether neural networks can be designed to learn an algorithm that scales with the square root of the problem size, or even logarithmically. 

\subsection{Even harder chess puzzles}
\label{sec:app_chess_harder}

\begin{figure}[h!]
    \centering
    \includegraphics[width=0.49\textwidth]{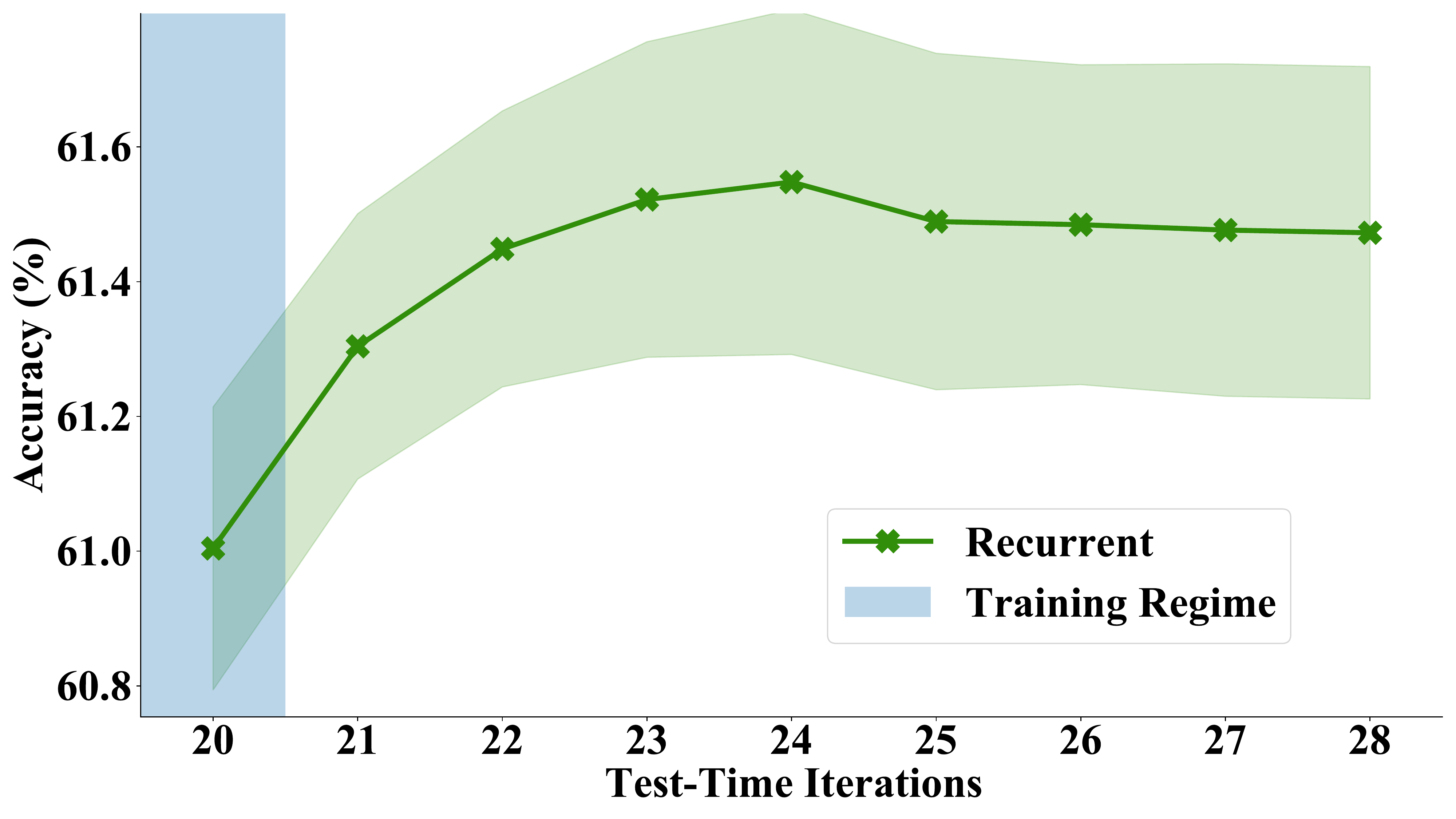}
    \includegraphics[width=0.49\textwidth]{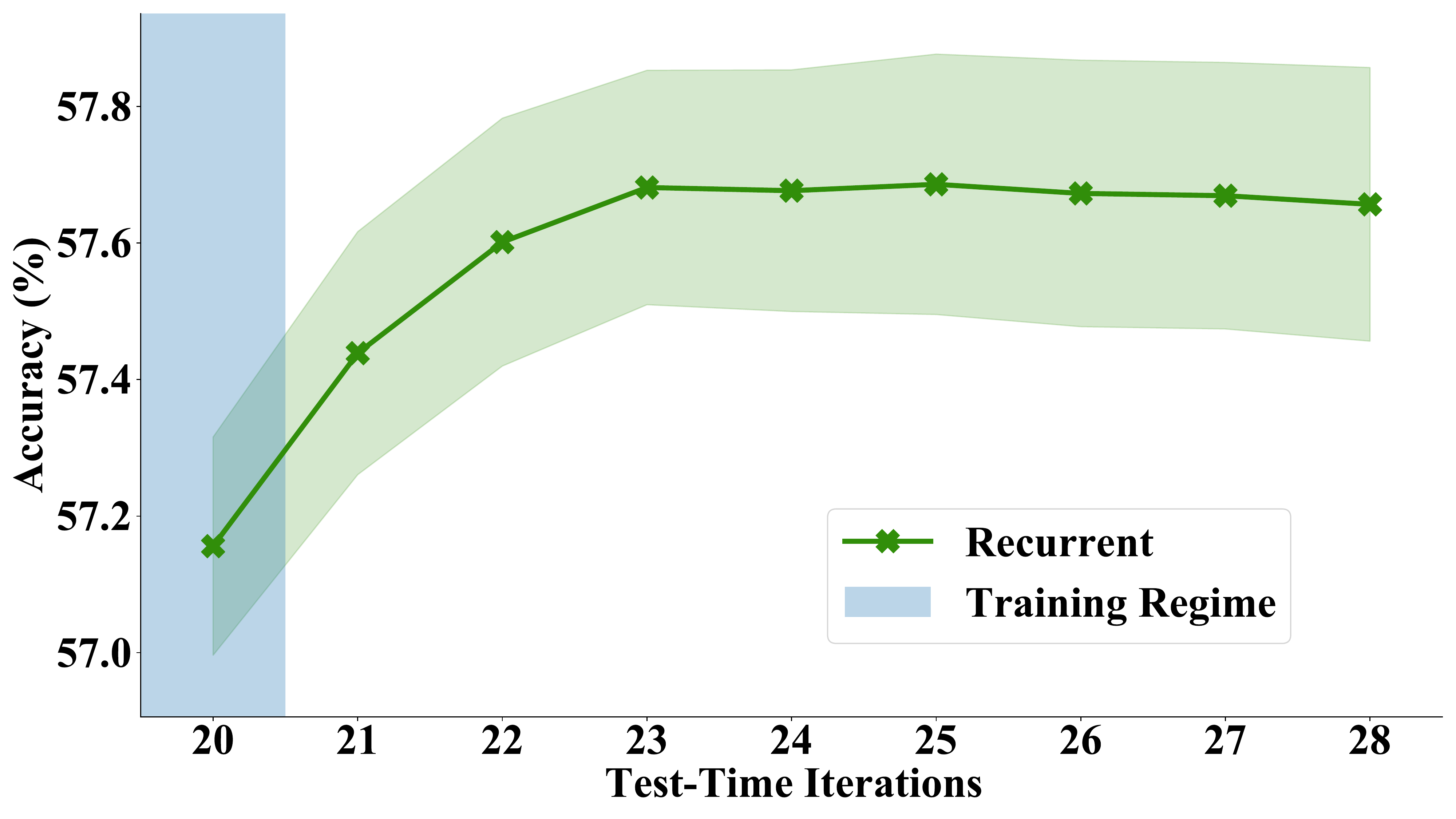}\\
    \includegraphics[width=0.49\textwidth]{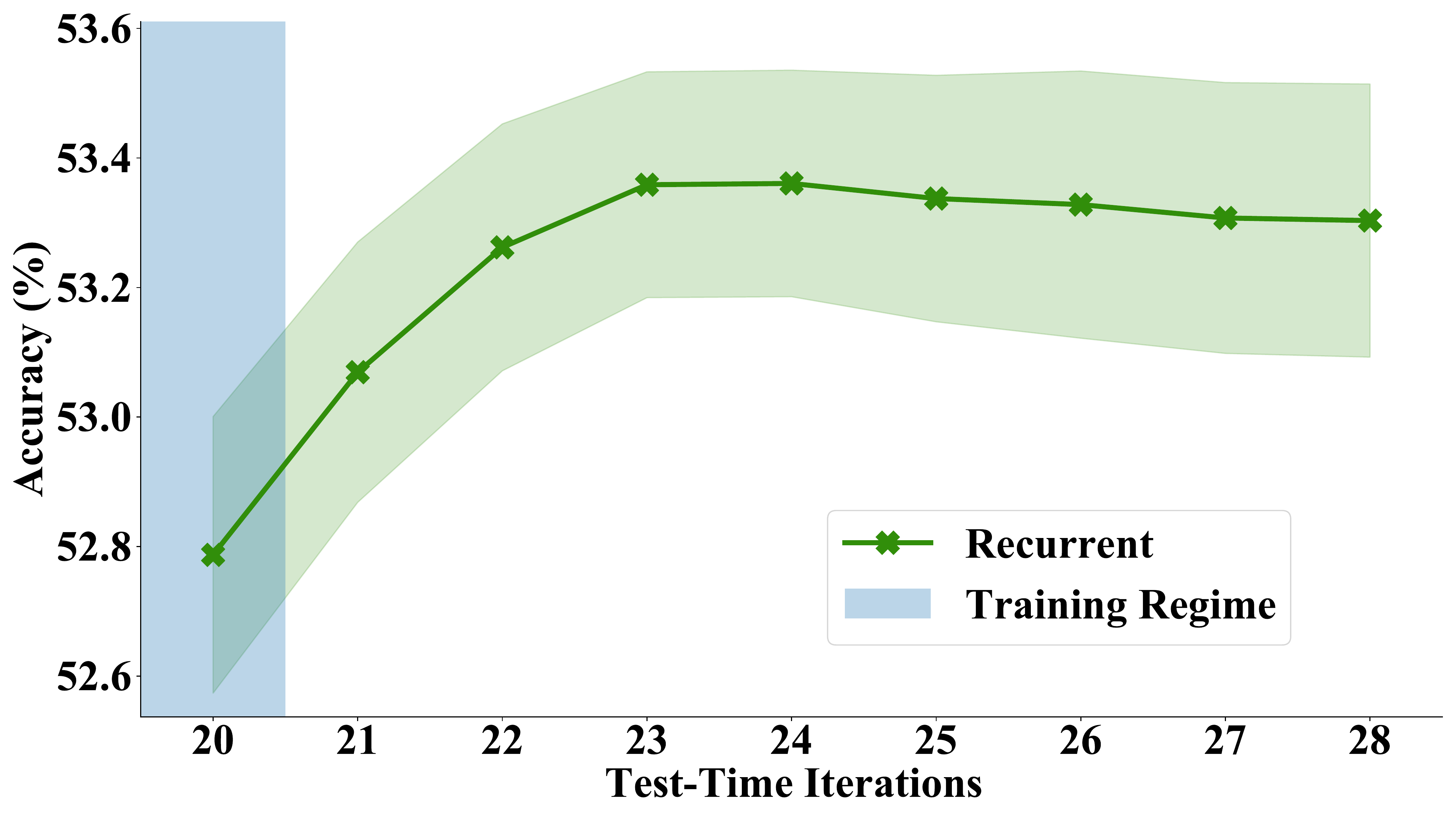}
    \includegraphics[width=0.49\textwidth]{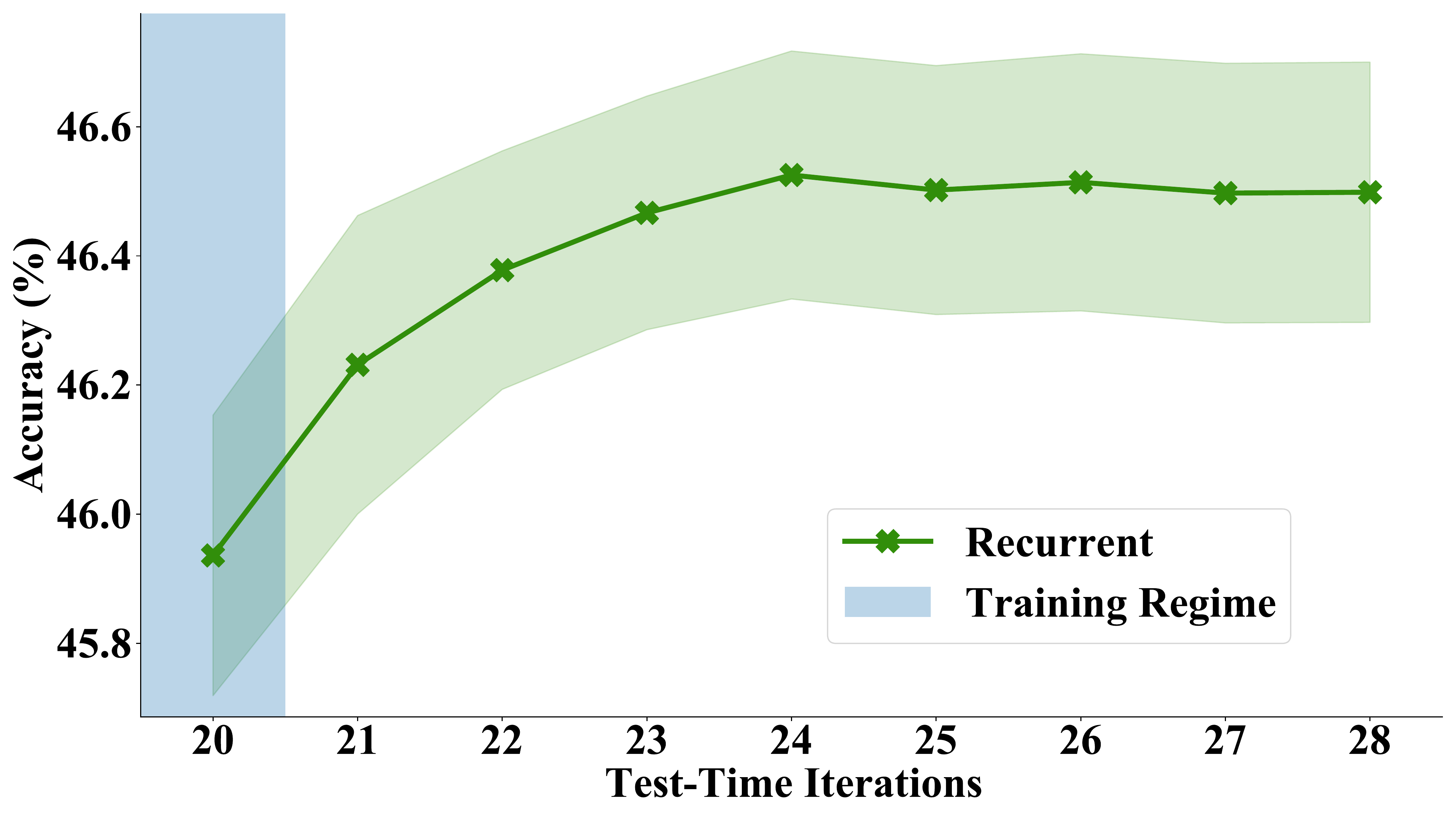}
    \caption{Generalizing from easy to hard chess puzzles. The ability of networks to solve harder puzzles than were used for training. We trained models on the first 600,000 puzzles and we show their performance with extra iterations on puzzles with index 800,000 to 850,000 (top left), 850,000 to 900,000 (top right), 900,000 to 950,000 (bottom left), and 100,000 to 150,000 (bottom right).  }
    \label{fig:app_chess_leap}
\end{figure}

\section{Visualizations}
\label{sec:app_visualizations}

Additional visualizations of intermediate outputs, along with input and target examples from all three datasets are presented below.

\subsection{Prefix sums}

We show a recurrent model's output from each of 11 iterations on 40-bit input strings. Shown below is the confidence that there is a 1 at each index of the output. The first index is at the top for all vectors, the input is in the left-most column and the  target is in the right-most column. The model used to produce these plots was trained with fewer iterations (10) on shorter input strings (32-bit).

\begin{figure}[h!]
    \centering
    \includegraphics[width=0.49\textwidth]{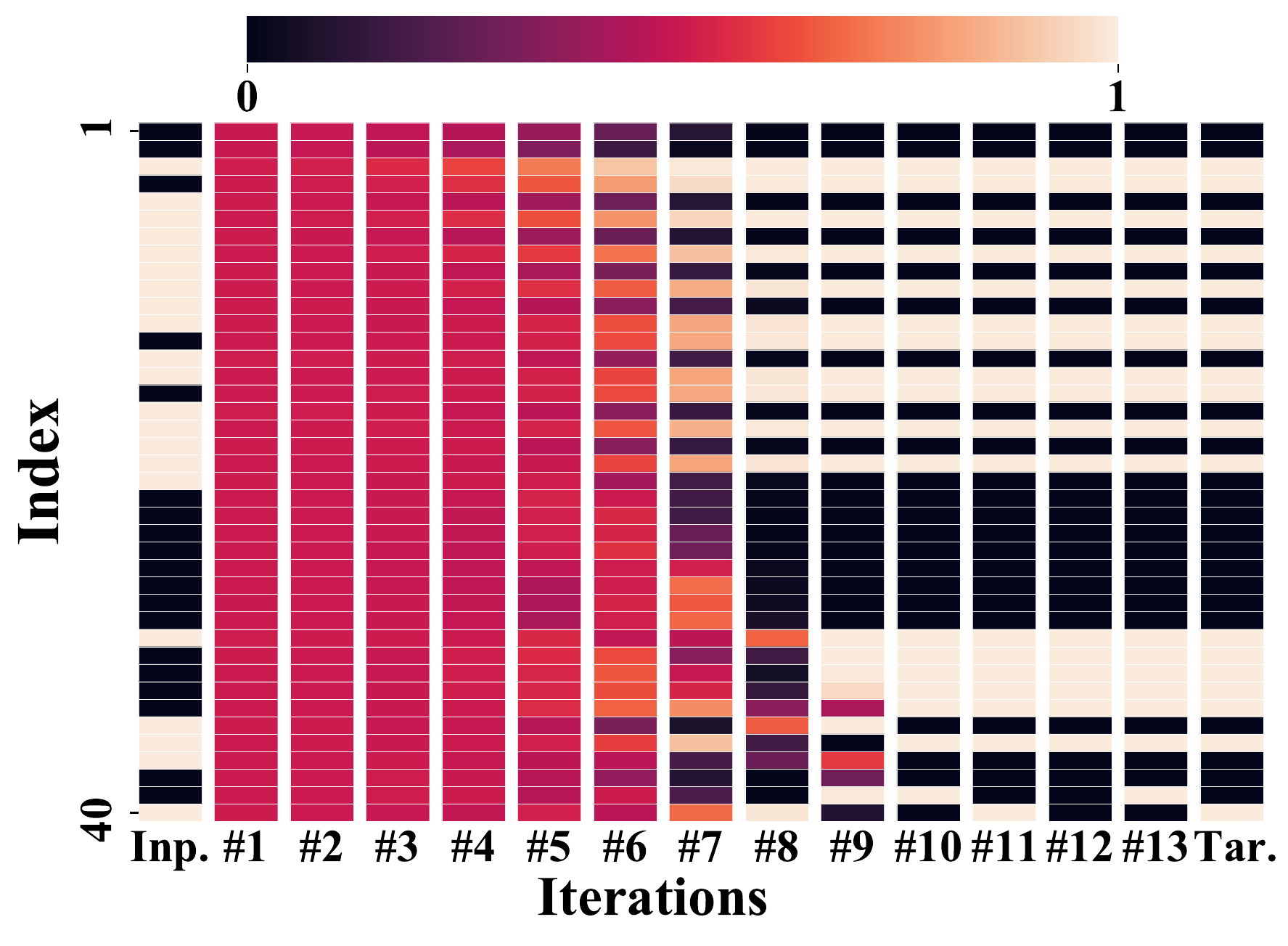}
    \includegraphics[width=0.49\textwidth]{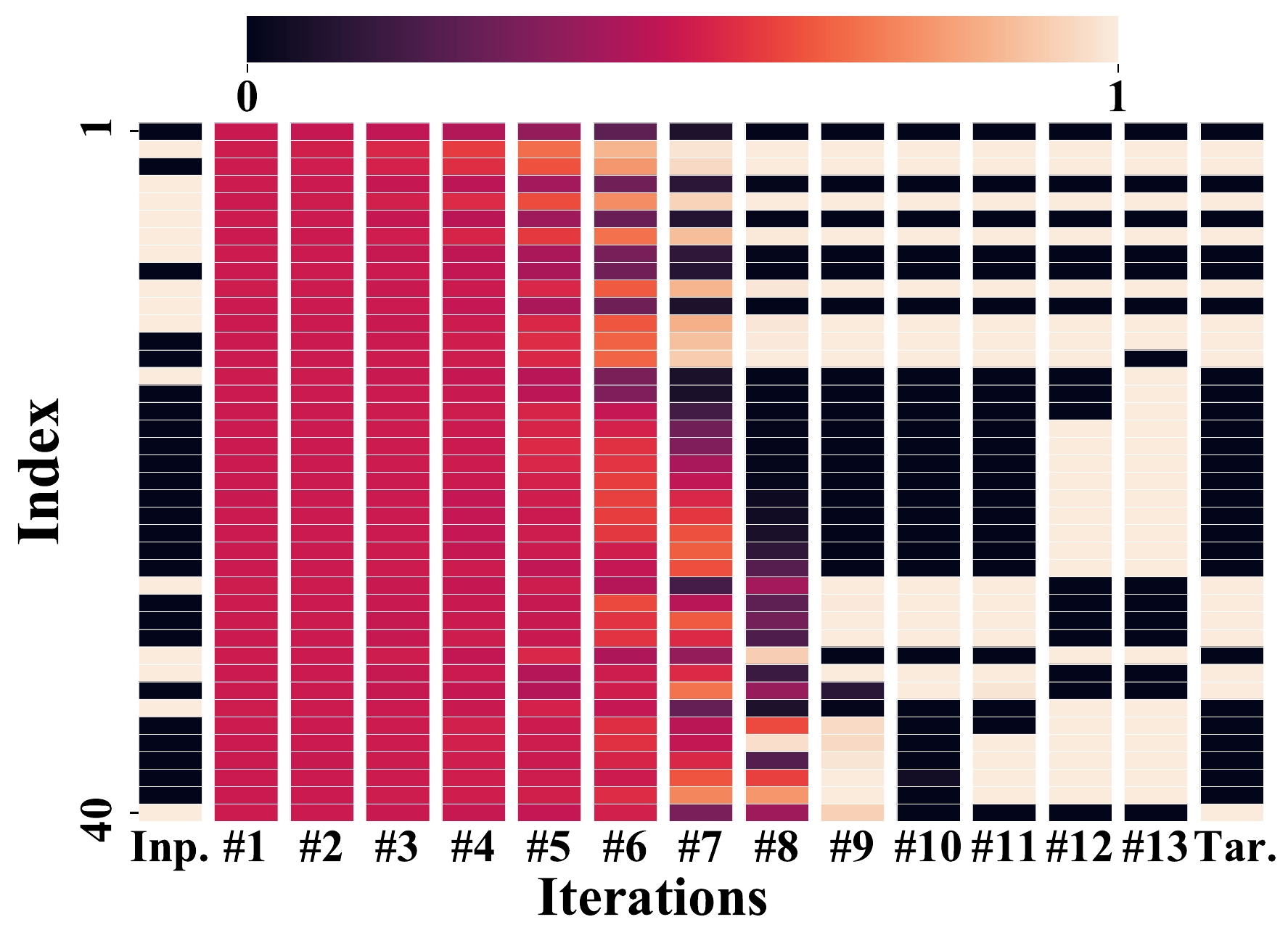} \\
    \includegraphics[width=0.49\textwidth]{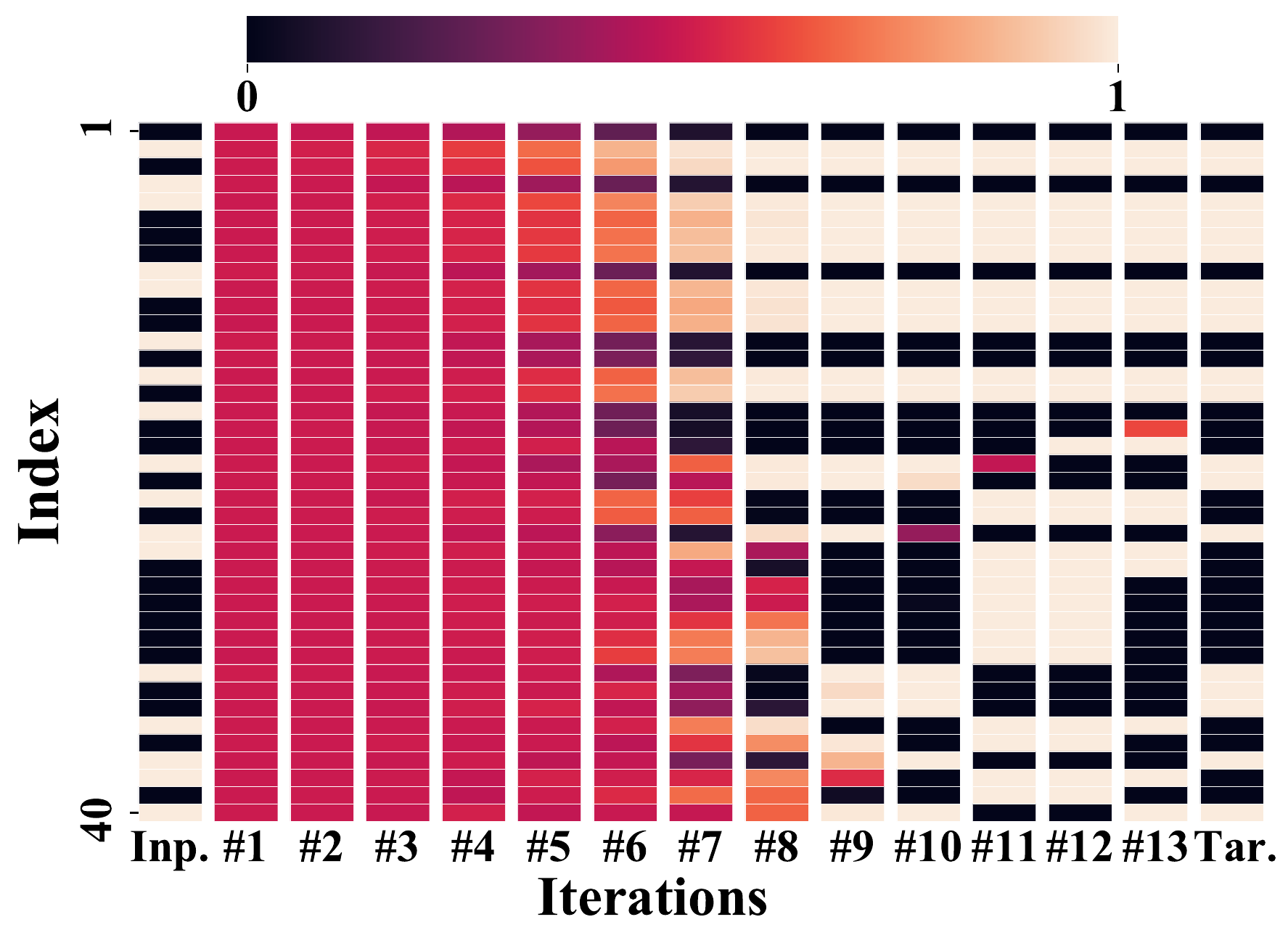}
    \includegraphics[width=0.49\textwidth]{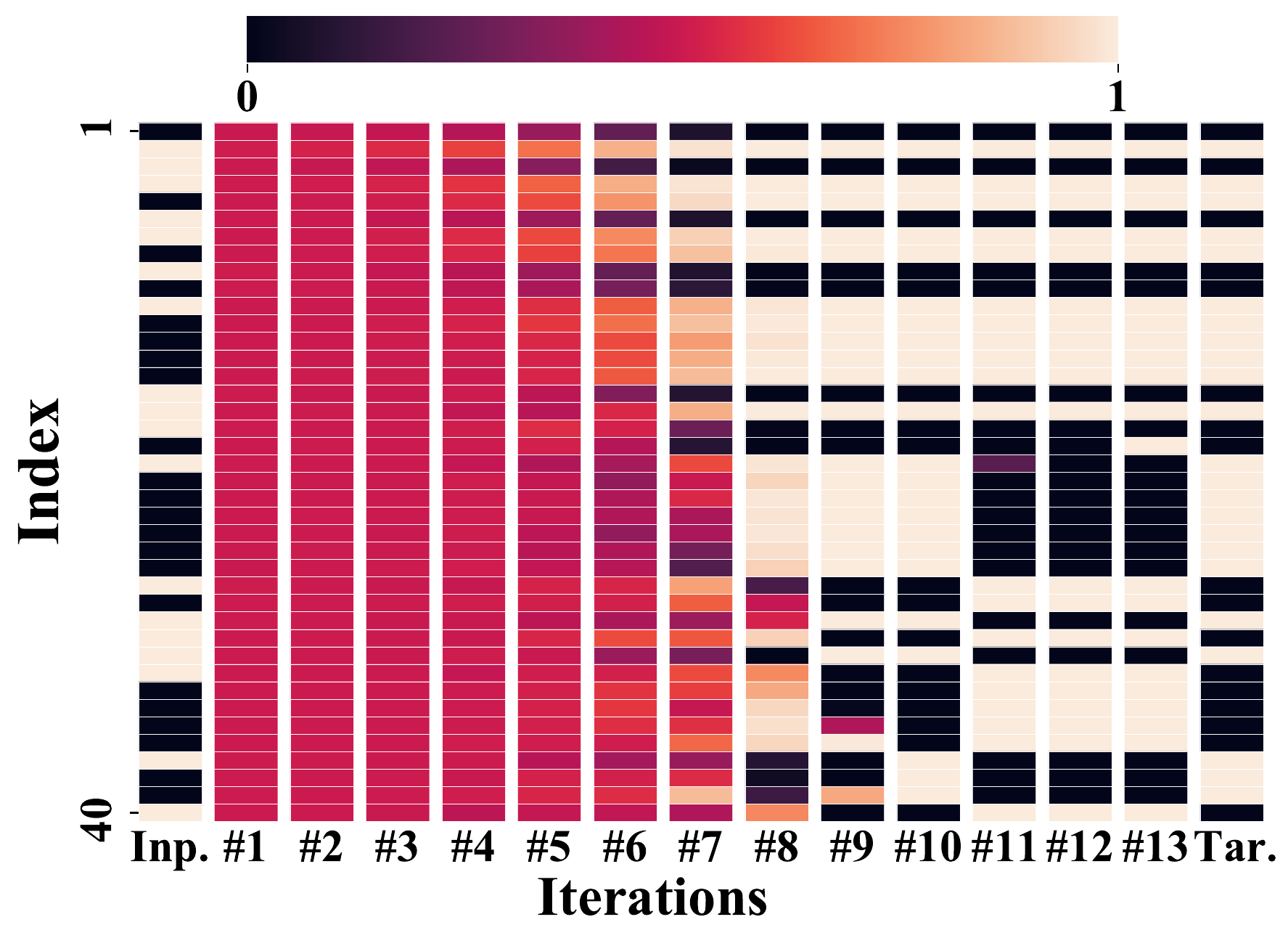}
    \caption{Prefix sum intermediate outputs.}
    \label{fig:app_prefix_thoughts}
\end{figure}

\subsection{Mazes}

We show inputs, targets, and outputs form different iterations to highlight the model's ability to think sequentially about mazes. We plot the model's confidence that each pixel belongs to the optimal path. Below are several representative examples from a model trained to solve small mazes in six iterations.

\begin{figure}[h!]
    \centering
    \raisebox{0.09\height}{\includegraphics[width=0.24\textwidth]{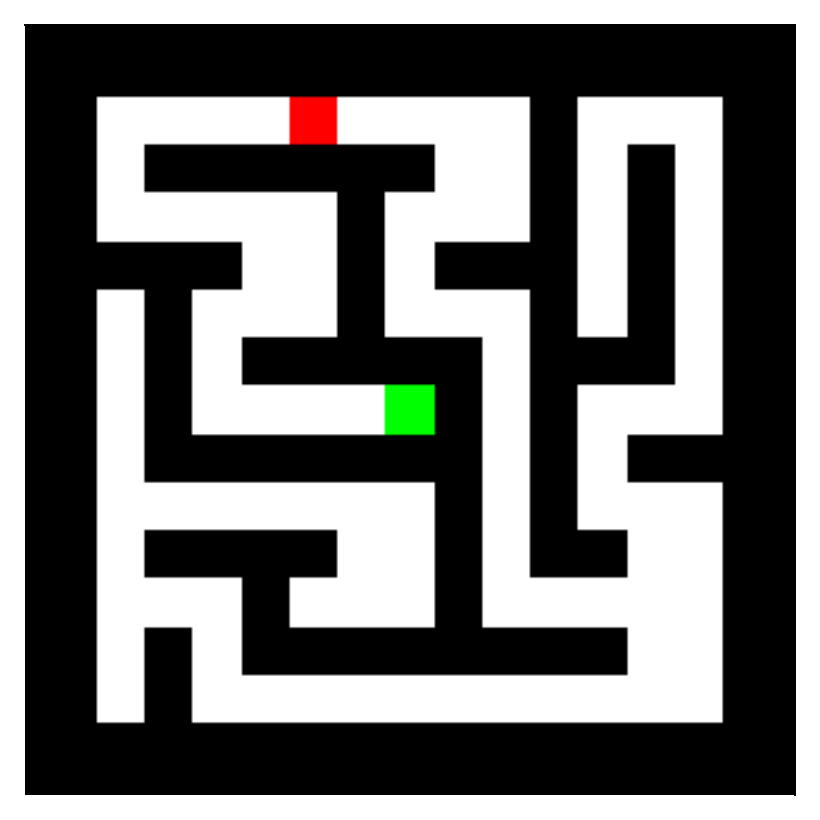}}
    \includegraphics[width=0.66\textwidth, trim=0cm 0.25cm 0cm 0cm, clip]{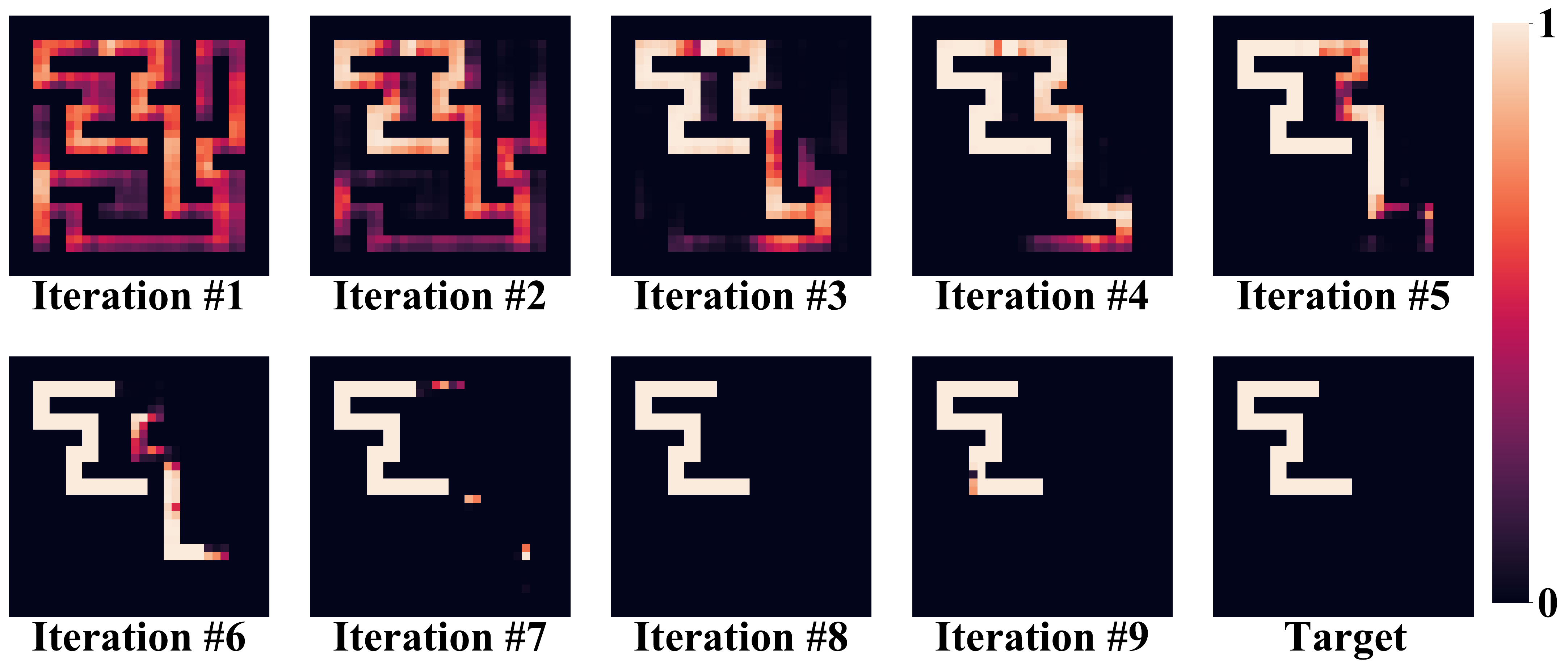} \\
        \raisebox{0.09\height}{\includegraphics[width=0.24\textwidth]{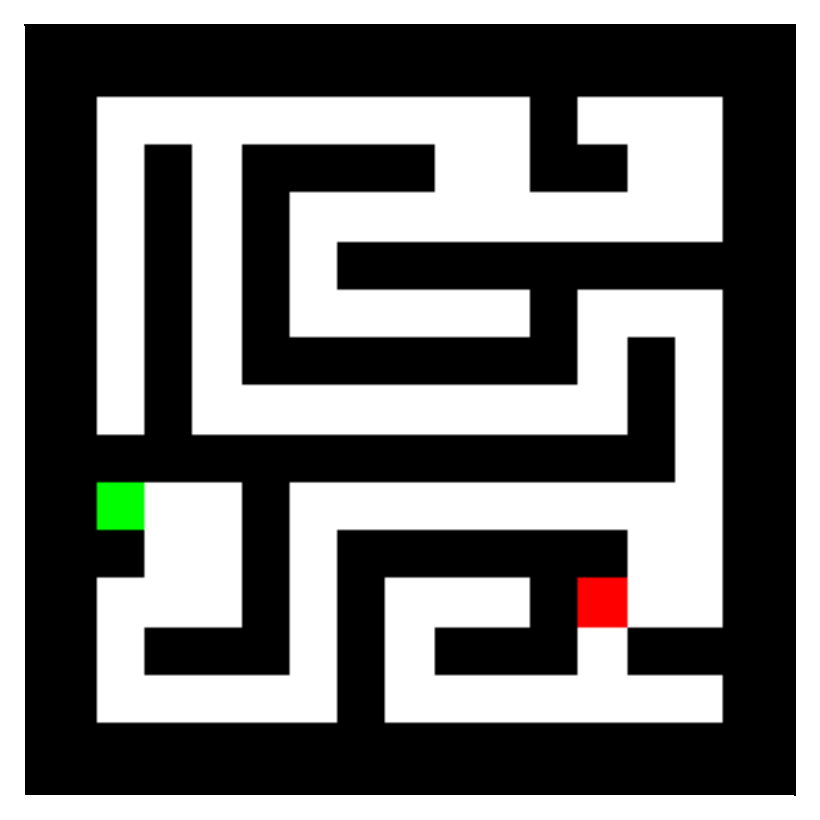}}
    \includegraphics[width=0.66\textwidth, trim=0cm 0.25cm 0cm 0cm, clip]{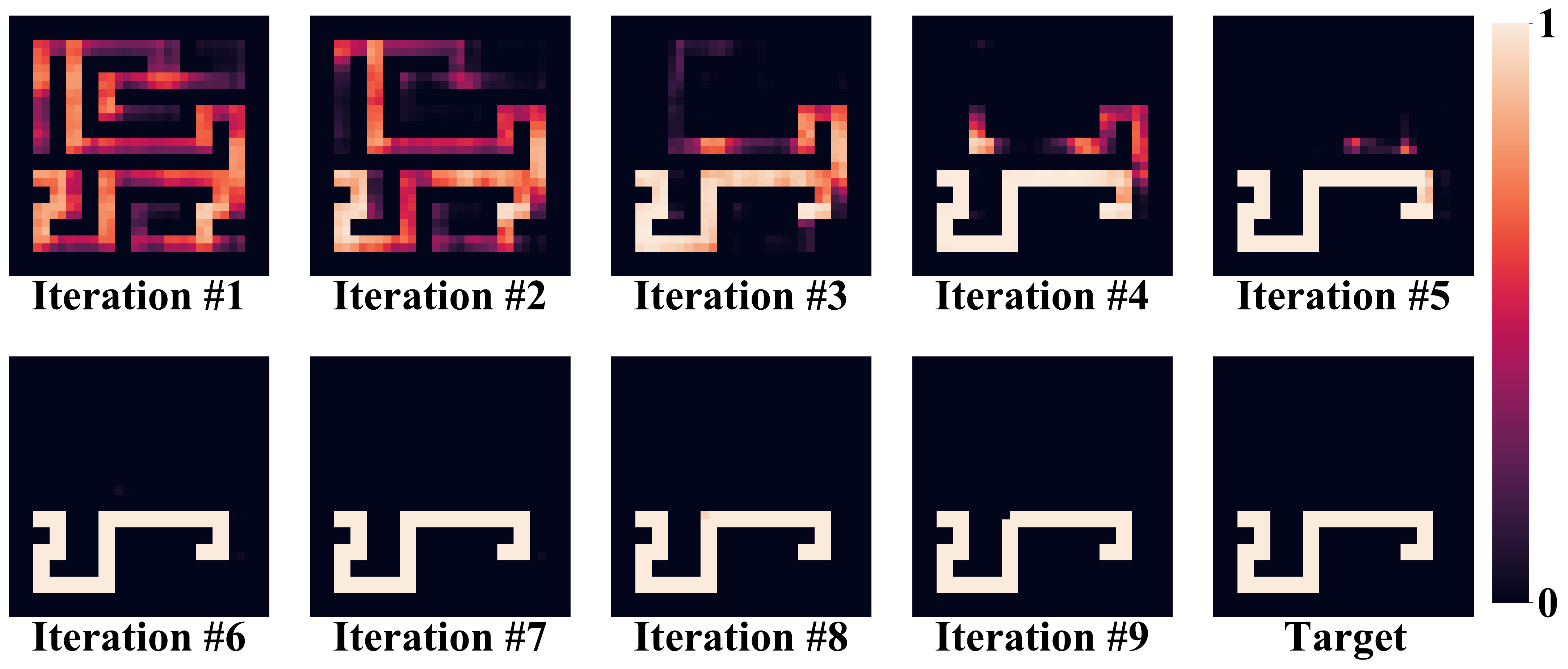} \\
        \raisebox{0.09\height}{\includegraphics[width=0.24\textwidth]{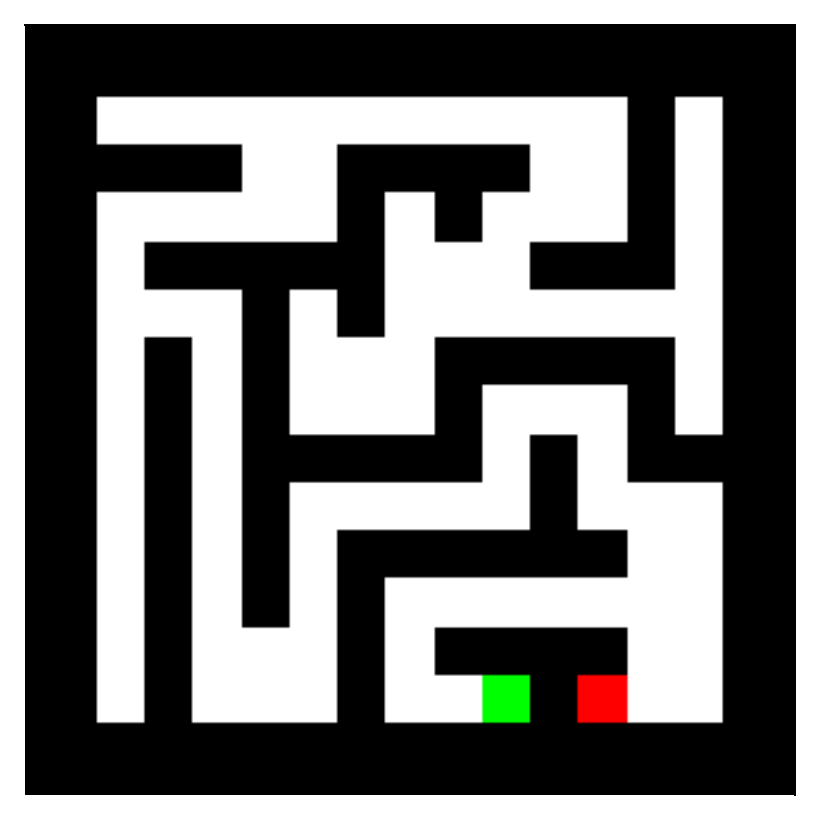}}
    \includegraphics[width=0.66\textwidth, trim=0cm 0.25cm 0cm 0cm, clip]{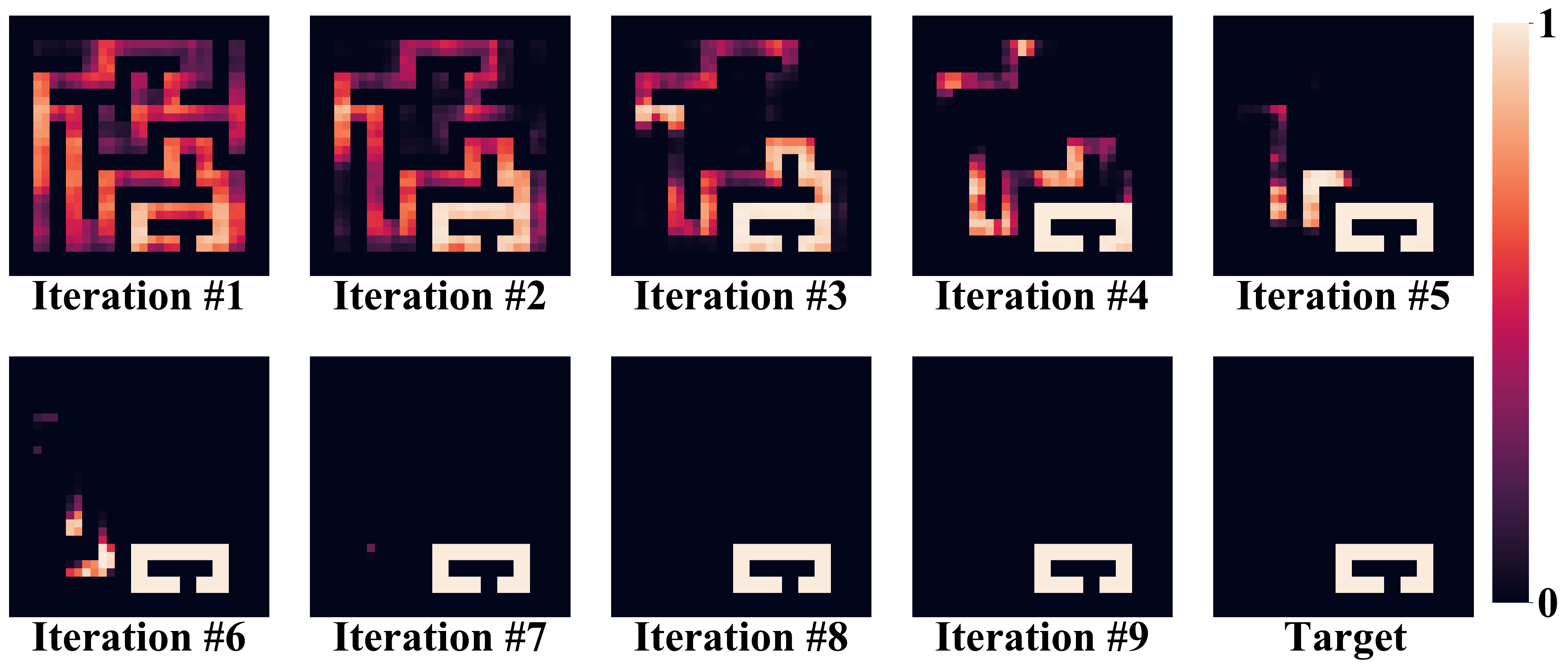} \\
        \raisebox{0.09\height}{\includegraphics[width=0.24\textwidth]{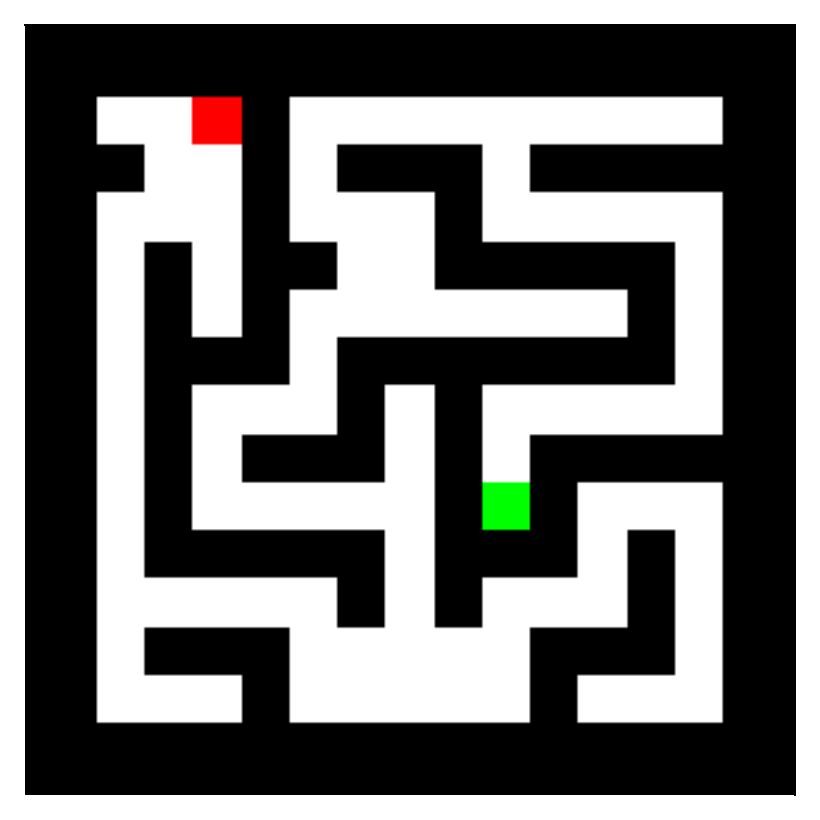}}
    \includegraphics[width=0.66\textwidth, trim=0cm 0.25cm 0cm 0cm, clip]{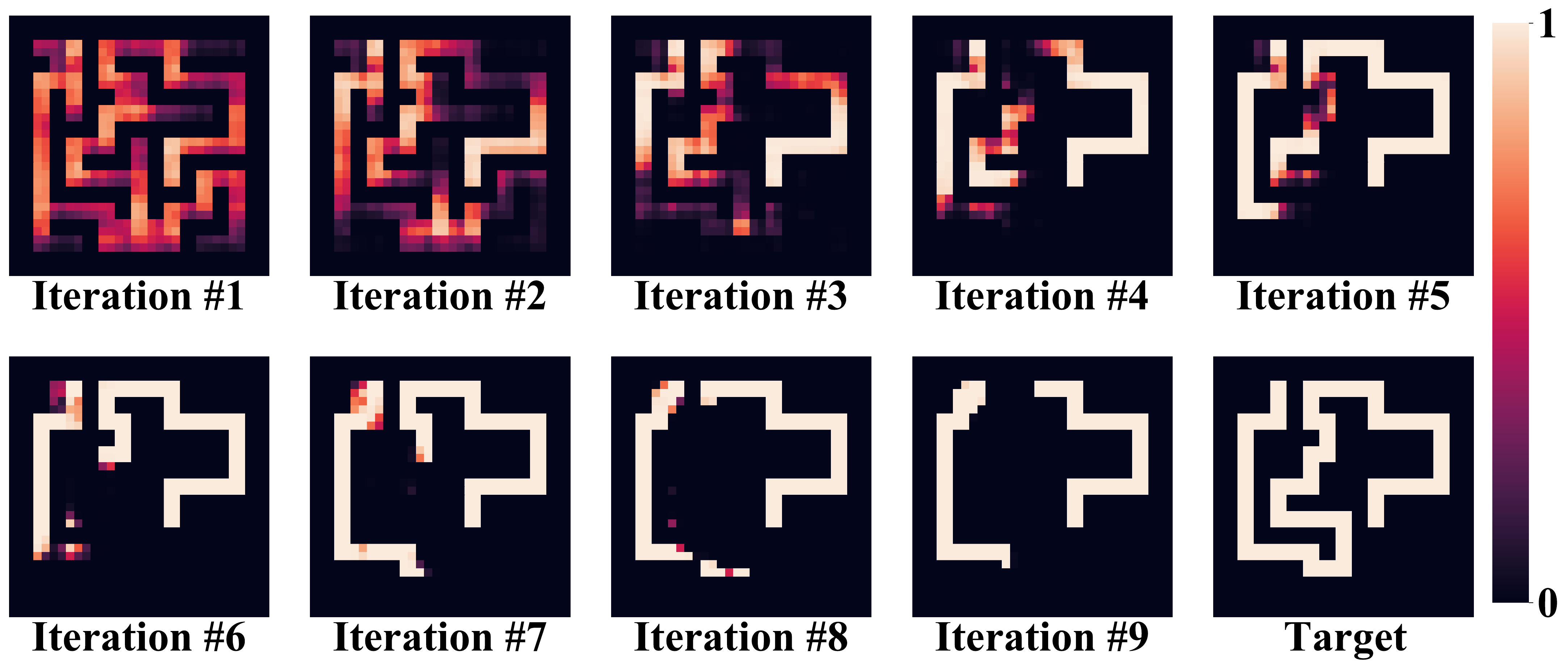}
    \caption{Maze model intermediate outputs.}
    \label{fig:app_mazes_thoughts}
\end{figure}

\subsection{Chess puzzles}

We show inputs, targets, and outputs form different iterations to highlight the model's ability to think about the next move. Below, we plot the model's confidence that each pixel is one of the two that define a move.

\begin{figure}[h!]
    \centering
    \raisebox{0.1\height}{\includegraphics[width=0.26\textwidth]{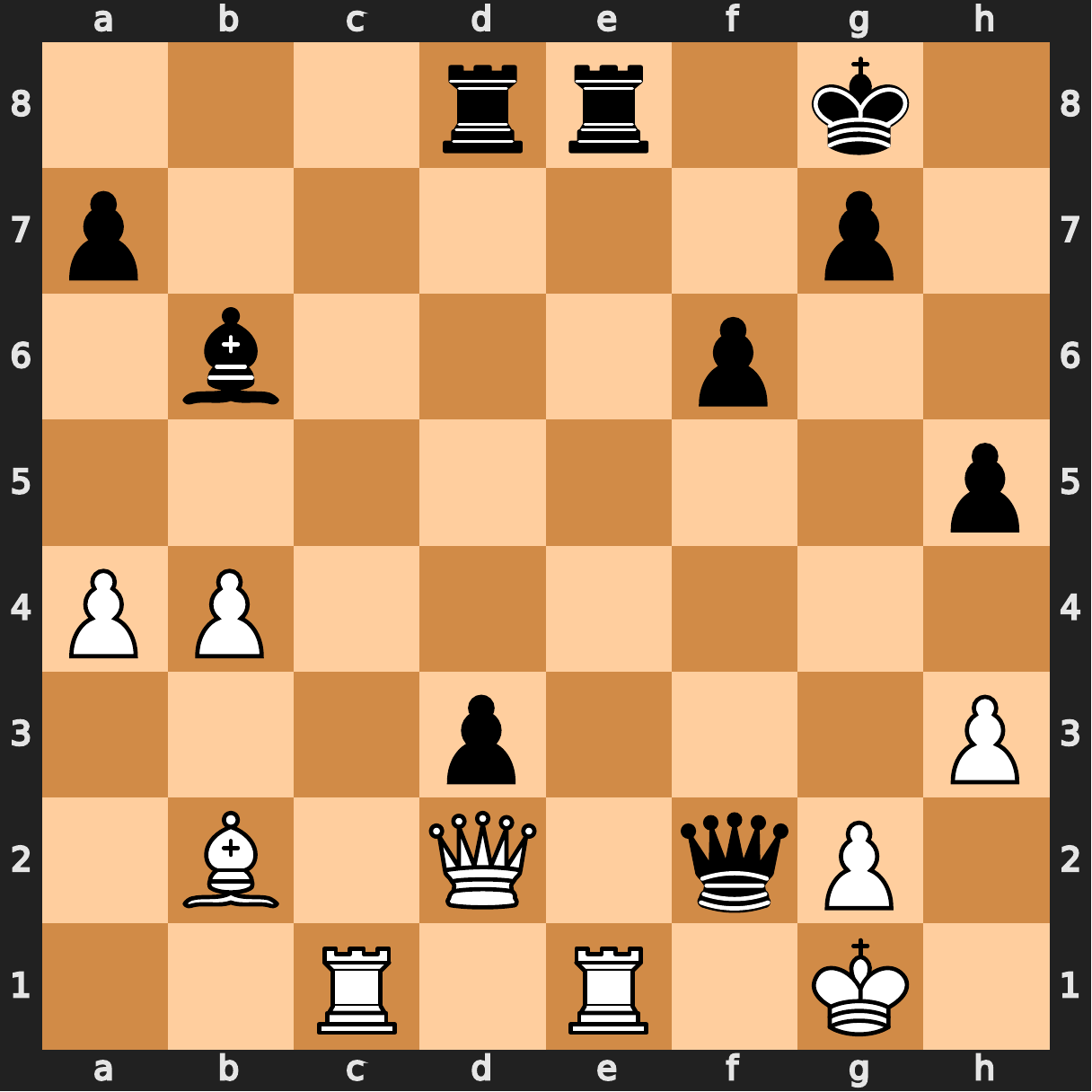}
    \includegraphics[width=0.64\textwidth, trim=0cm 0.25cm 0cm 0cm, clip]{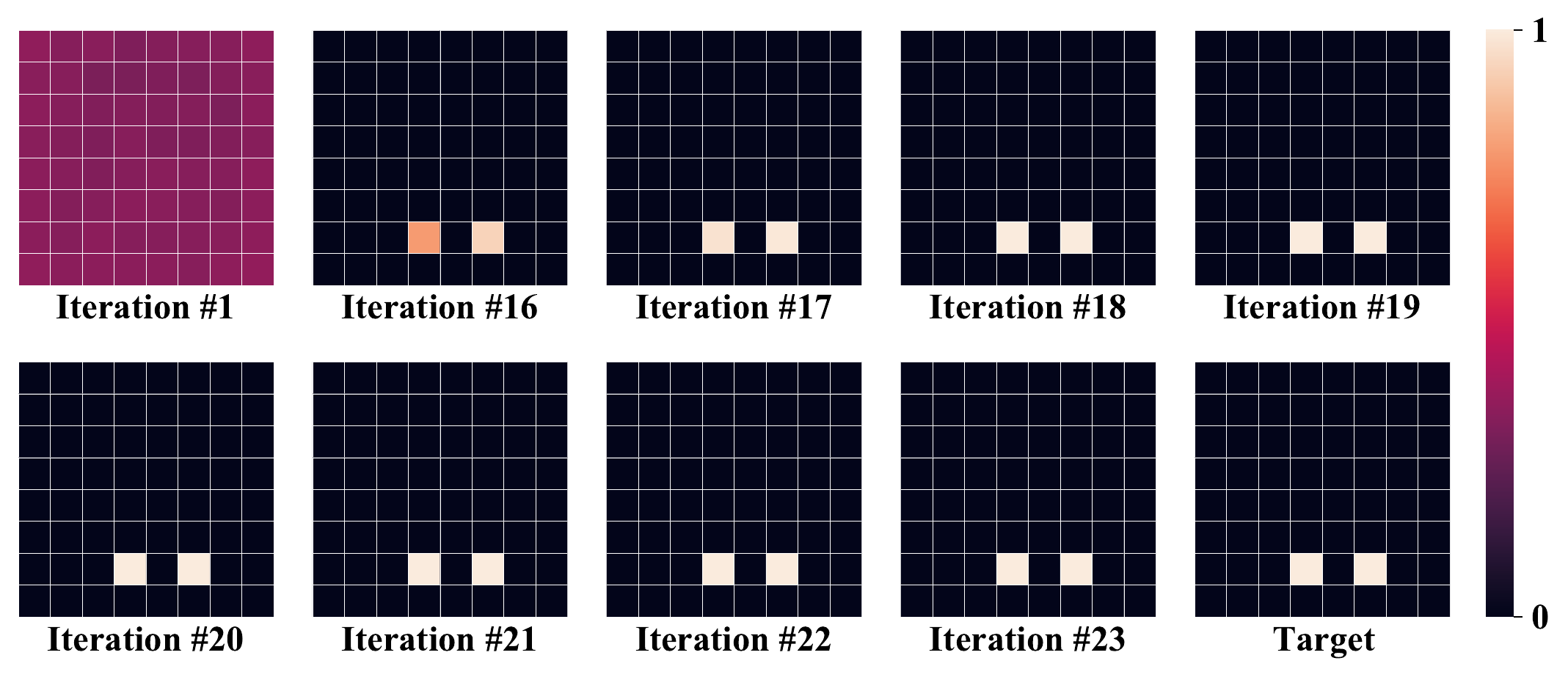}} \\
    \raisebox{0.1\height}{\includegraphics[width=0.26\textwidth]{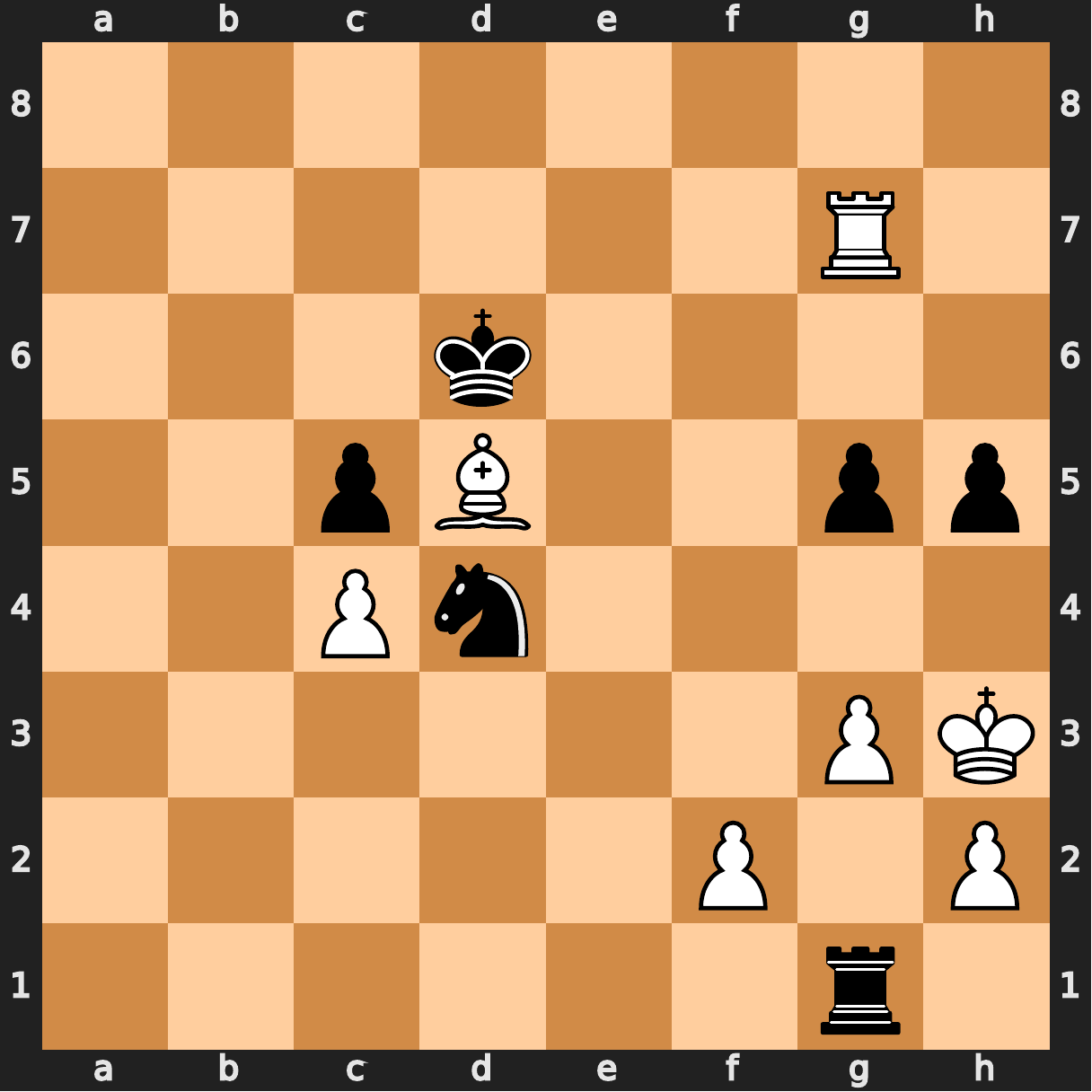}
    \includegraphics[width=0.64\textwidth, trim=0cm 0.25cm 0cm 0cm, clip]{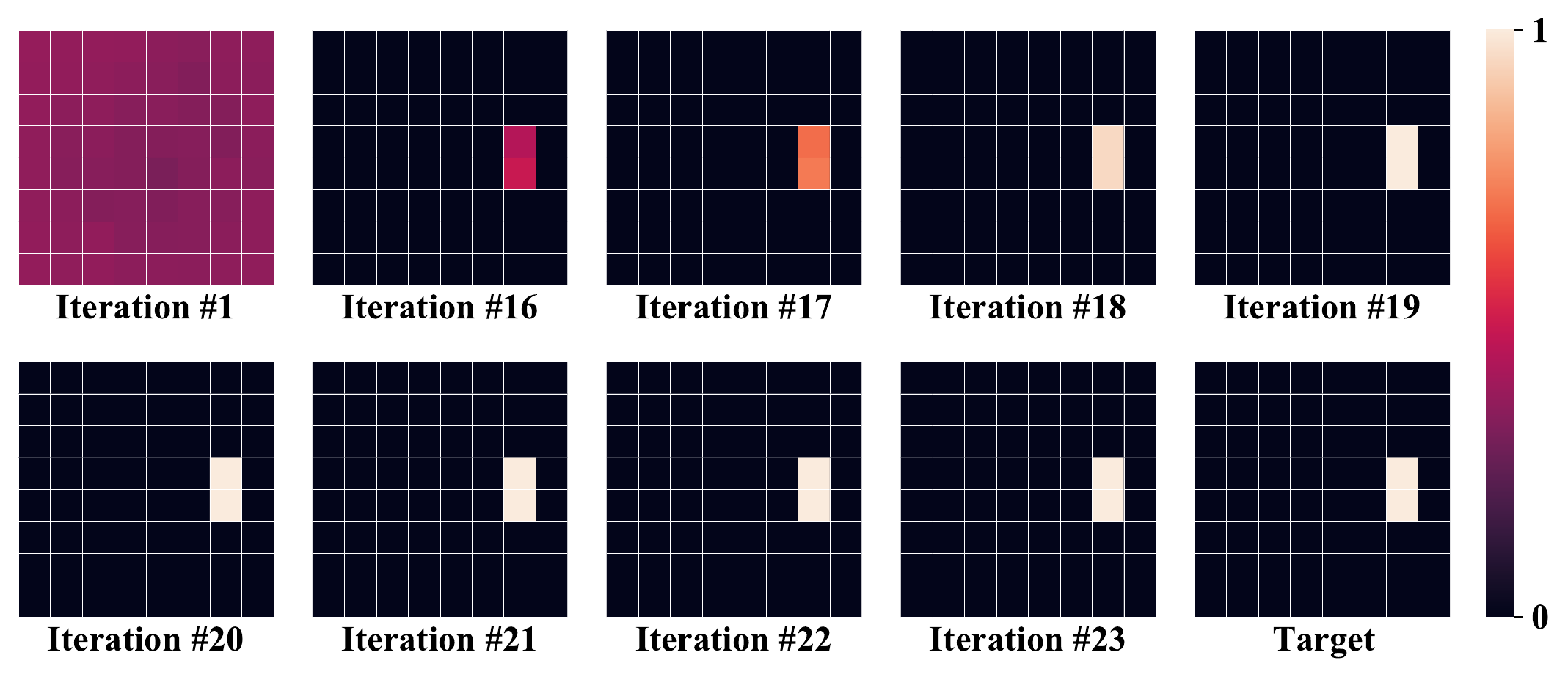}} \\
    \raisebox{0.1\height}{\includegraphics[width=0.26\textwidth]{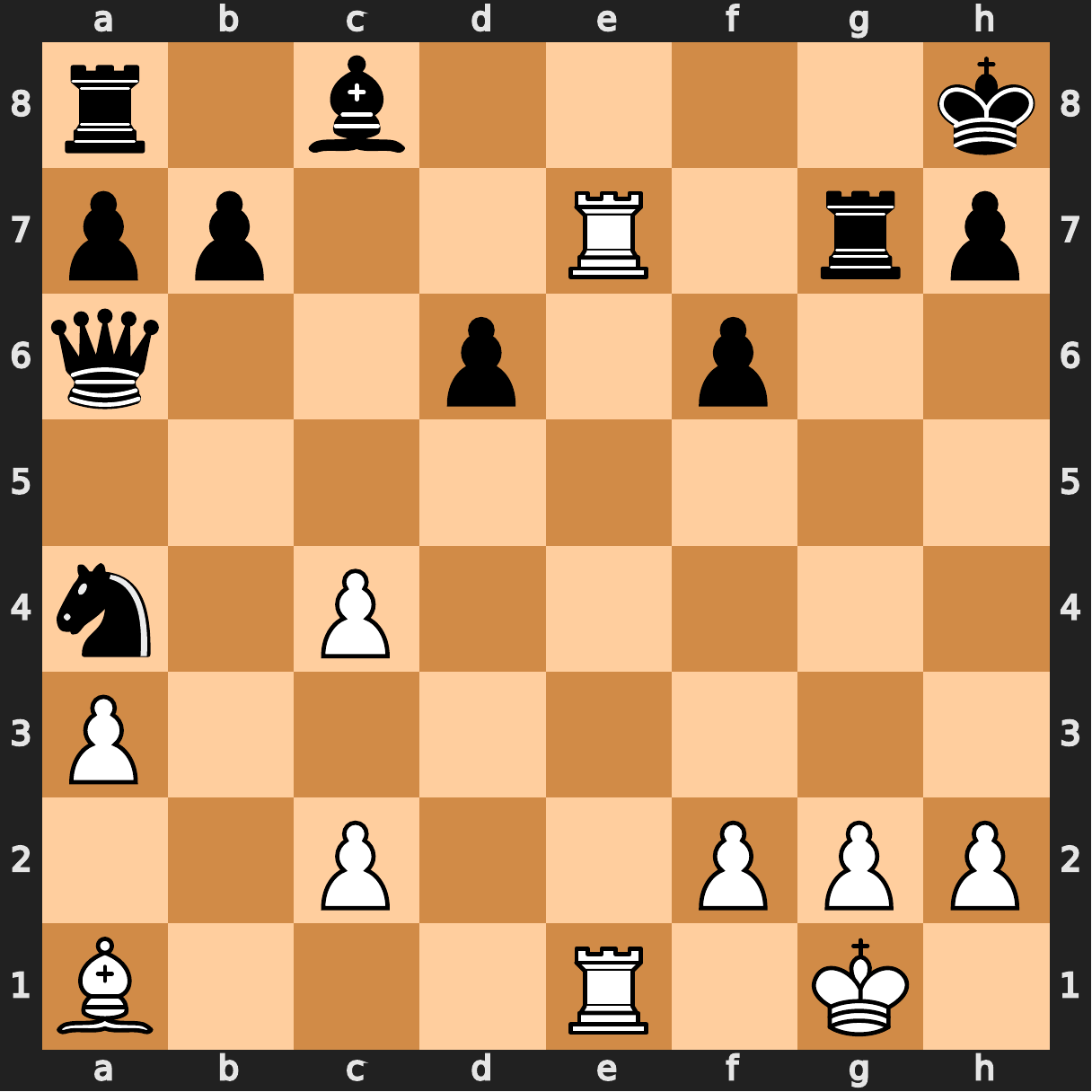}
    \includegraphics[width=0.64\textwidth, trim=0cm 0.25cm 0cm 0cm, clip]{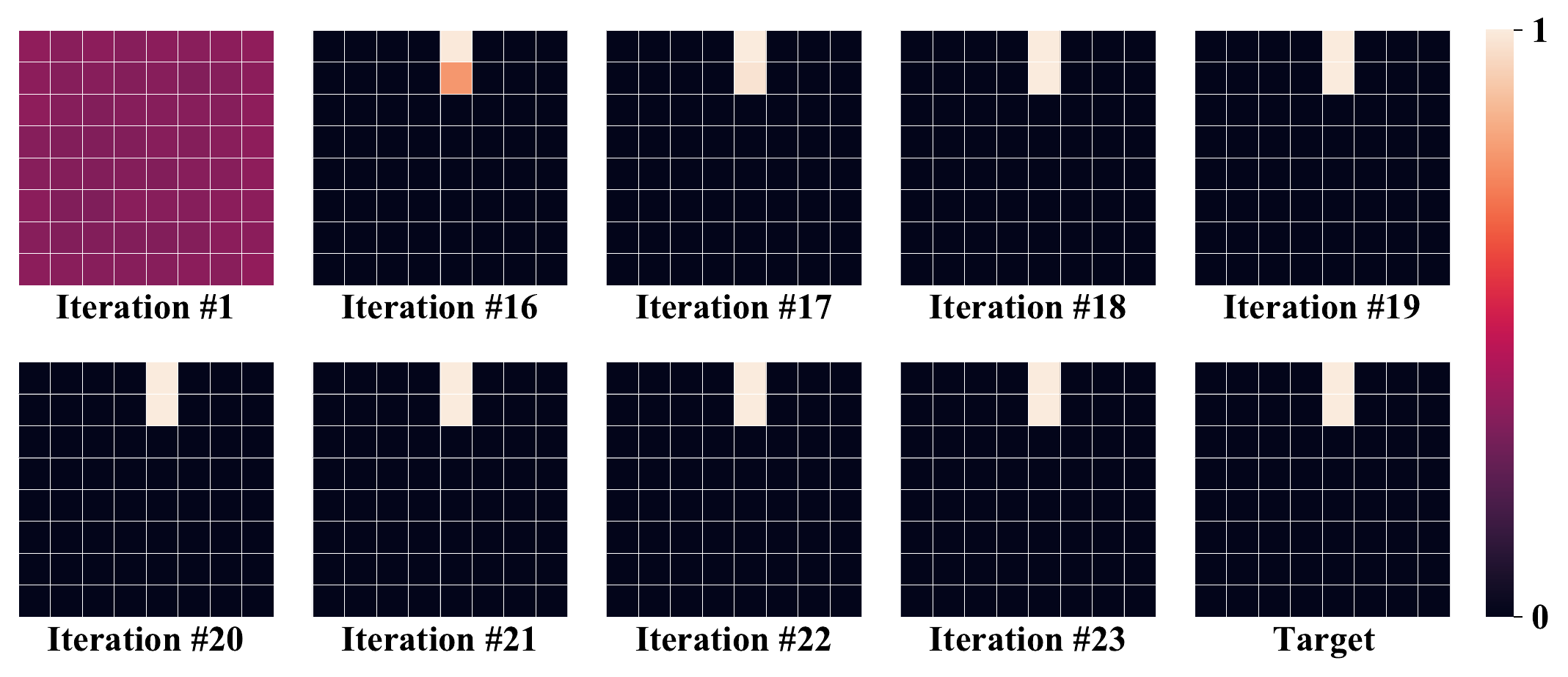}} \\
    \raisebox{0.1\height}{\includegraphics[width=0.26\textwidth]{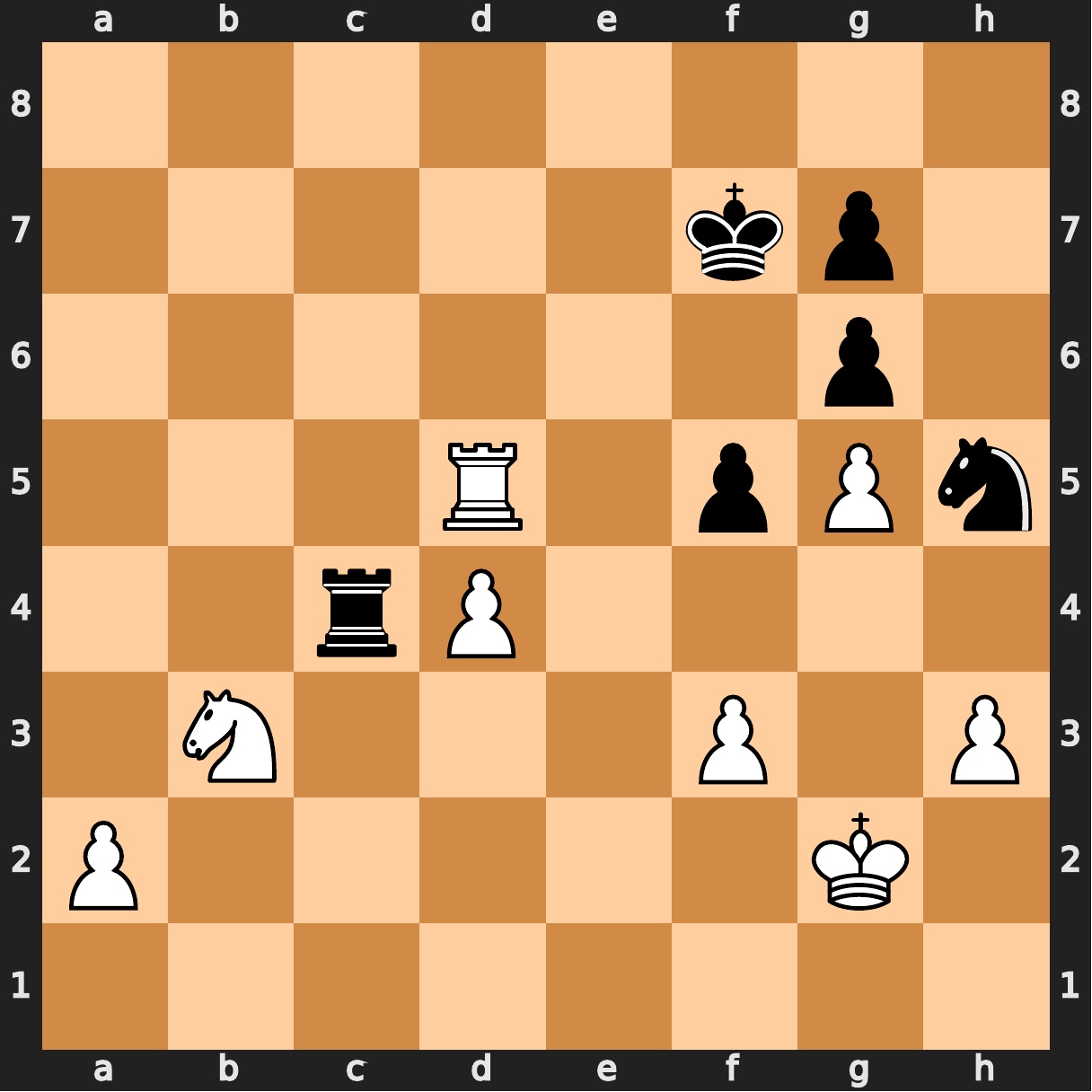}
    \includegraphics[width=0.64\textwidth, trim=0cm 0.25cm 0cm 0cm, clip]{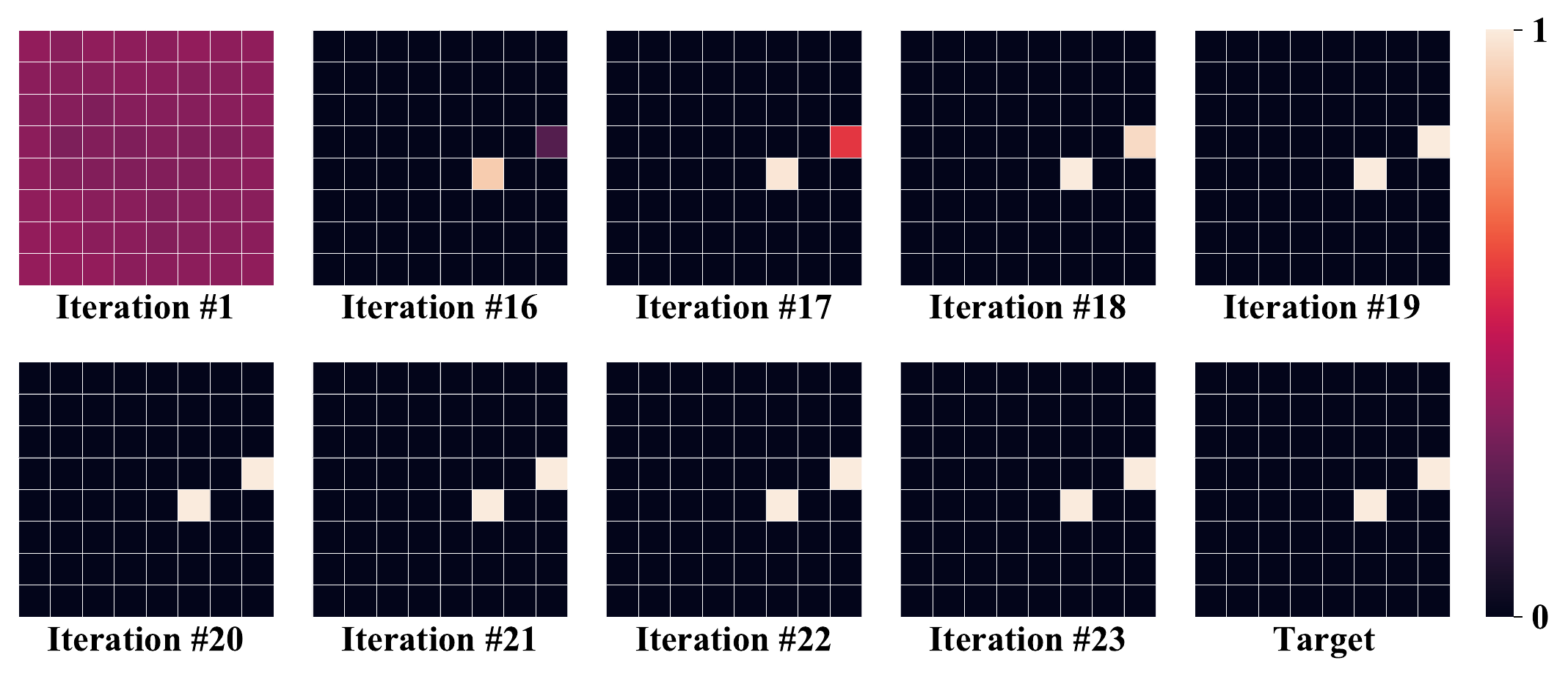}} 
    \caption{Chess model intermediate outputs.}
    \label{fig:app_chess_thoughts}
\end{figure}

\clearpage 

\section{NeurIPS submission checklist}
\begin{enumerate}

\item For all authors...
\begin{enumerate}
  \item Do the main claims made in the abstract and introduction accurately reflect the paper's contributions and scope?
    \answerYes{}
  \item Did you describe the limitations of your work?
    \answerYes{See Section \ref{sec:discussion}.}
  \item Did you discuss any potential negative societal impacts of your work?
    \answerNo{We feel that our investigation into the behavior of AI in some toy domains does not warrant such a discussion, as it would relate only to the potential of the field in general.}
  \item Have you read the ethics review guidelines and ensured that your paper conforms to them?
    \answerYes{}
\end{enumerate}

\item If you are including theoretical results...
\begin{enumerate}
  \item Did you state the full set of assumptions of all theoretical results?
    \answerNA{}
	\item Did you include complete proofs of all theoretical results?
    \answerNA{}
\end{enumerate}

\item If you ran experiments...
\begin{enumerate}
  \item Did you include the code, data, and instructions needed to reproduce the main experimental results (either in the supplemental material or as a URL)?
    \answerYes{See Appendix Sections \ref{sec:app_datasets}, \ref{sec:app_architectures}, and \ref{sec:app_hyperparams}, as well as the code in the supplementary files.}
  \item Did you specify all the training details (e.g., data splits, hyperparameters, how they were chosen)?
    \answerYes{See Appendix \ref{sec:app_hyperparams}.}
	\item Did you report error bars (e.g., with respect to the random seed after running experiments multiple times)?
    \answerYes{}
	\item Did you include the total amount of compute and the type of resources used (e.g., type of GPUs, internal cluster, or cloud provider)?
    \answerYes{See Appendix \ref{sec:app_compute}.}
\end{enumerate}

\item If you are using existing assets (e.g., code, data, models) or curating/releasing new assets...
\begin{enumerate}
  \item If your work uses existing assets, did you cite the creators?
    \answerYes{See Section \ref{sec:data}.}
  \item Did you mention the license of the assets?
    \answerYes{See Appendix \ref{sec:app_chess_data}.}
  \item Did you include any new assets either in the supplemental material or as a URL?
    \answerYes{See Appendix \ref{sec:app_datasets}.}
  \item Did you discuss whether and how consent was obtained from people whose data you're using/curating?
    \answerYes{Lichess is open source and encourages the use of their data for research projects, see \url{https://database.lichess.org/}. This is mentioned in Section \ref{sec:data}.}
  \item Did you discuss whether the data you are using/curating contains personally identifiable information or offensive content?
    \answerYes{See Appendix \ref{sec:app_chess_data}.}
\end{enumerate}

\item If you used crowdsourcing or conducted research with human subjects...
\begin{enumerate}
  \item Did you include the full text of instructions given to participants and screenshots, if applicable?
    \answerNA{}
  \item Did you describe any potential participant risks, with links to Institutional Review Board (IRB) approvals, if applicable?
    \answerNA{}
  \item Did you include the estimated hourly wage paid to participants and the total amount spent on participant compensation?
    \answerNA{}
\end{enumerate}

\end{enumerate}

%% file: main.bbl
\begin{thebibliography}{30}
\providecommand{\natexlab}[1]{#1}
\providecommand{\url}[1]{\texttt{#1}}
\expandafter\ifx\csname urlstyle\endcsname\relax
  \providecommand{\doi}[1]{doi: #1}\else
  \providecommand{\doi}{doi: \begingroup \urlstyle{rm}\Url}\fi

\bibitem[Alom et~al.(2018)Alom, Hasan, Yakopcic, Taha, and
  Asari]{alom2018recurrent}
Md~Zahangir Alom, Mahmudul Hasan, Chris Yakopcic, Tarek~M Taha, and Vijayan~K
  Asari.
\newblock Recurrent residual convolutional neural network based on u-net
  (r2u-net) for medical image segmentation.
\newblock \emph{arXiv preprint arXiv:1802.06955}, 2018.

\bibitem[Arjovsky et~al.(2019)Arjovsky, Bottou, Gulrajani, and
  Lopez-Paz]{arjovsky2019invariant}
Martin Arjovsky, L{\'e}on Bottou, Ishaan Gulrajani, and David Lopez-Paz.
\newblock Invariant risk minimization.
\newblock \emph{arXiv preprint arXiv:1907.02893}, 2019.

\bibitem[Baddeley(2012)]{baddeley2012working}
Alan Baddeley.
\newblock Working memory: theories, models, and controversies.
\newblock \emph{Annual review of psychology}, 63:\penalty0 1--29, 2012.

\bibitem[Baddeley and Hitch(1974)]{baddeley1974working}
Alan~D Baddeley and Graham Hitch.
\newblock Working memory.
\newblock In \emph{Psychology of learning and motivation}, volume~8, pages
  47--89. Elsevier, 1974.

\bibitem[Bai et~al.(2018)Bai, Kolter, and Koltun]{bai2018trellis}
Shaojie Bai, J~Zico Kolter, and Vladlen Koltun.
\newblock Trellis networks for sequence modeling.
\newblock In \emph{International Conference on Learning Representations}, 2018.

\bibitem[Bai et~al.(2019)Bai, Kolter, and Koltun]{bai2019deep}
Shaojie Bai, J~Zico Kolter, and Vladlen Koltun.
\newblock Deep equilibrium models.
\newblock \emph{Advances in Neural Information Processing Systems},
  32:\penalty0 690--701, 2019.

\bibitem[Biswas and Regan(2015)]{biswas2015measuring}
Tamal Biswas and Kenneth Regan.
\newblock Measuring level-k reasoning, satisficing, and human error in
  game-play data.
\newblock In \emph{2015 IEEE 14th International Conference on Machine Learning
  and Applications (ICMLA)}, pages 941--947. IEEE, 2015.

\bibitem[Elo(1978)]{elo1978rating}
Arpad~E Elo.
\newblock \emph{The rating of chessplayers, past and present}.
\newblock Arco Pub., 1978.

\bibitem[Eyzaguirre and Soto(2020)]{eyzaguirre2020differentiable}
Cristobal Eyzaguirre and Alvaro Soto.
\newblock Differentiable adaptive computation time for visual reasoning.
\newblock In \emph{Proceedings of the IEEE/CVF Conference on Computer Vision
  and Pattern Recognition}, pages 12817--12825, 2020.

\bibitem[Graves(2016)]{graves2016adaptive}
Alex Graves.
\newblock Adaptive computation time for recurrent neural networks.
\newblock \emph{arXiv preprint arXiv:1603.08983}, 2016.

\bibitem[Graves et~al.(2014)Graves, Wayne, and Danihelka]{graves2014neural}
Alex Graves, Greg Wayne, and Ivo Danihelka.
\newblock Neural turing machines.
\newblock \emph{arXiv preprint arXiv:1410.5401}, 2014.

\bibitem[{He} et~al.(2016){He}, {Zhang}, {Ren}, and {Sun}]{he2016deep}
K.~{He}, X.~{Zhang}, S.~{Ren}, and J.~{Sun}.
\newblock Deep residual learning for image recognition.
\newblock In \emph{2016 IEEE Conference on Computer Vision and Pattern
  Recognition (CVPR)}, pages 770--778, 2016.

\bibitem[Hill(2017)]{hill2017making}
Christian Hill.
\newblock Making a maze, Apr 2017.
\newblock URL \url{https://scipython.com/blog/making-a-maze/}.

\bibitem[Huang et~al.(2016)Huang, Sun, Liu, Sedra, and
  Weinberger]{huang2016deep}
Gao Huang, Yu~Sun, Zhuang Liu, Daniel Sedra, and Kilian~Q Weinberger.
\newblock Deep networks with stochastic depth.
\newblock In \emph{European conference on computer vision}, pages 646--661.
  Springer, 2016.

\bibitem[Jaegle et~al.(2021)Jaegle, Gimeno, Brock, Zisserman, Vinyals, and
  Carreira]{jaegle2021perceiver}
Andrew Jaegle, Felix Gimeno, Andrew Brock, Andrew Zisserman, Oriol Vinyals, and
  Joao Carreira.
\newblock Perceiver: General perception with iterative attention.
\newblock \emph{arXiv preprint arXiv:2103.03206}, 2021.

\bibitem[Kaiser and Sutskever(2015)]{kaiser2015neural}
{\L}ukasz Kaiser and Ilya Sutskever.
\newblock Neural gpus learn algorithms.
\newblock \emph{arXiv preprint arXiv:1511.08228}, 2015.

\bibitem[Kar et~al.(2019)Kar, Kubilius, Schmidt, Issa, and
  DiCarlo]{kar2019evidence}
Kohitij Kar, Jonas Kubilius, Kailyn Schmidt, Elias~B Issa, and James~J DiCarlo.
\newblock Evidence that recurrent circuits are critical to the ventral
  stream’s execution of core object recognition behavior.
\newblock \emph{Nature neuroscience}, 22\penalty0 (6):\penalty0 974--983, 2019.

\bibitem[Kaya et~al.(2019)Kaya, Hong, and Dumitras]{kaya2019shallow}
Yigitcan Kaya, Sanghyun Hong, and Tudor Dumitras.
\newblock Shallow-deep networks: Understanding and mitigating network
  overthinking.
\newblock In \emph{International Conference on Machine Learning}, pages
  3301--3310. PMLR, 2019.

\bibitem[Lan et~al.(2020)Lan, Chen, Goodman, Gimpel, Sharma, and
  Soricut]{lan2020albert}
Zhenzhong Lan, Mingda Chen, Sebastian Goodman, Kevin Gimpel, Piyush Sharma, and
  Radu Soricut.
\newblock Albert: A lite bert for self-supervised learning of language
  representations.
\newblock In \emph{International Conference on Learning Representations}, 2020.

\bibitem[Liang and Hu(2015)]{liang2015recurrent}
Ming Liang and Xiaolin Hu.
\newblock Recurrent convolutional neural network for object recognition.
\newblock In \emph{Proceedings of the IEEE conference on computer vision and
  pattern recognition}, pages 3367--3375, 2015.

\bibitem[Liao and Poggio(2016)]{liao2016bridging}
Qianli Liao and Tomaso Poggio.
\newblock Bridging the gaps between residual learning, recurrent neural
  networks and visual cortex.
\newblock \emph{arXiv preprint arXiv:1604.03640}, 2016.

\bibitem[Lichess(2021)]{lichess}
Lichess.
\newblock Lichess open puzzles database.
\newblock \url{https://database.lichess.org/#puzzles}, 2021.
\newblock Accessed: 2021-04-01.

\bibitem[McIlroy-Young et~al.(2020)McIlroy-Young, Sen, Kleinberg, and
  Anderson]{mcilroy2020aligning}
Reid McIlroy-Young, Siddhartha Sen, Jon Kleinberg, and Ashton Anderson.
\newblock Aligning superhuman ai with human behavior: Chess as a model system.
\newblock In \emph{Proceedings of the 26th ACM SIGKDD International Conference
  on Knowledge Discovery \& Data Mining}, pages 1677--1687, 2020.

\bibitem[Pinheiro and Collobert(2014)]{pinheiro2014recurrent}
Pedro Pinheiro and Ronan Collobert.
\newblock Recurrent convolutional neural networks for scene labeling.
\newblock In \emph{International conference on machine learning}, pages 82--90.
  PMLR, 2014.

\bibitem[Romstad et~al.()Romstad, Costalba, and Kiiski]{stockfish}
Tord Romstad, Marco Costalba, and et~al. Kiiski, Joona.
\newblock Stockfish: A strong open source chess engine.
\newblock URL \url{https://stockfishchess.org/}.

\bibitem[Schwarzschild et~al.(2021)Schwarzschild, Borgnia, Gupta, Bansal, Emam,
  Huang, Goldblum, and Goldstein]{schwarzschild2021datasets}
Avi Schwarzschild, Eitan Borgnia, Arjun Gupta, Arpit Bansal, Zeyad Emam, Furong
  Huang, Micah Goldblum, and Tom Goldstein.
\newblock Datasets for studying generalization from easy to hard examples.
\newblock \emph{arXiv preprint arXiv:2108.06011}, 2021.

\bibitem[Selsam et~al.(2018)Selsam, Lamm, B{\"u}nz, Liang, de~Moura, and
  Dill]{selsam2018learning}
Daniel Selsam, Matthew Lamm, Benedikt B{\"u}nz, Percy Liang, Leonardo de~Moura,
  and David~L Dill.
\newblock Learning a sat solver from single-bit supervision.
\newblock \emph{arXiv preprint arXiv:1802.03685}, 2018.

\bibitem[Shu et~al.(2020)Shu, Wu, Goldblum, and Goldstein]{shu2020preparing}
Manli Shu, Zuxuan Wu, Micah Goldblum, and Tom Goldstein.
\newblock Preparing for the worst: Making networks less brittle with
  adversarial batch normalization.
\newblock \emph{arXiv preprint arXiv:2009.08965}, 2020.

\bibitem[Silver et~al.(2017)Silver, Hubert, Schrittwieser, Antonoglou, Lai,
  Guez, Lanctot, Sifre, Kumaran, Graepel, et~al.]{silver2017mastering}
David Silver, Thomas Hubert, Julian Schrittwieser, Ioannis Antonoglou, Matthew
  Lai, Arthur Guez, Marc Lanctot, Laurent Sifre, Dharshan Kumaran, Thore
  Graepel, et~al.
\newblock Mastering chess and shogi by self-play with a general reinforcement
  learning algorithm.
\newblock \emph{arXiv preprint arXiv:1712.01815}, 2017.

\bibitem[Tamar et~al.(2016)Tamar, Wu, Thomas, Levine, and
  Abbeel]{tamar2016value}
Aviv Tamar, Yi~Wu, Garrett Thomas, Sergey Levine, and Pieter Abbeel.
\newblock Value iteration networks.
\newblock \emph{arXiv preprint arXiv:1602.02867}, 2016.

\end{thebibliography}
